\documentclass[sigconf,nonacm]{acmart}
\AtBeginDocument{%
  }

\setcopyright{acmlicensed}
\copyrightyear{2018}
\acmYear{2018}
\acmDOI{XXXXXXX.XXXXXXX}
\acmConference[Conference acronym 'XX]{Make sure to enter the correct
  conference title from your rights confirmation email}{June 03--05,
  2018}{Woodstock, NY}
\acmISBN{978-1-4503-XXXX-X/2018/06}
\usepackage{soul} 
\usepackage{colortbl}
\usepackage{tabularx,booktabs,array}
\usepackage{listings}
\usepackage[normalem]{ulem}
\usepackage{graphicx}
\usepackage{multirow}

\begin{document}

\title{Are LLMs Better GNN Helpers? Rethinking Robust Graph Learning under Deficiencies with Iterative Refinement}

\author{Zhaoyan Wang}
\authornote{Corresponding author.}
\affiliation{%
  \institution{School of Computing\\ KAIST}
  \city{Daejeon}
  \country{Republic of Korea}
}
\email{zhaoyan123@kaist.ac.kr}

\author{Zheng Gao}
\affiliation{%
  \institution{School of Computer Science and Engineering\\ UNSW}
  \city{Sydney}
  \state{NSW}
  \country{Australia}
}
\email{zheng.gao1@unsw.edu.au}

\author{Arogya Kharel}
\affiliation{%
  \institution{School of Computing\\ KAIST}
  \city{Daejeon}
  \country{Republic of Korea}
}
\email{akharel@kaist.ac.kr}

\author{In-Young Ko}
\affiliation{%
  \institution{School of Computing\\ KAIST}
  \city{Daejeon}
  \country{Republic of Korea}
}
\email{iko@kaist.ac.kr}

\renewcommand{\shortauthors}{Zhaoyan Wang et al.}

\begin{abstract}
Graph Neural Networks (GNNs) are widely adopted in Web-related applications, serving as a core technique for learning from graph-structured data, such as text-attributed graphs. Yet in real-world scenarios, such graphs exhibit deficiencies that substantially undermine GNN performance. While prior GNN-based augmentation studies have explored robustness against individual imperfections, a systematic understanding of how graph-native and Large Language Models (LLMs) enhanced methods behave under compound deficiencies is still missing. Specifically, there has been no comprehensive investigation comparing conventional approaches and recent LLM-on-graph frameworks, leaving their merits unclear. To fill this gap, we conduct the first empirical study that benchmarks these two lines of methods across diverse graph deficiencies, revealing overlooked vulnerabilities and challenging the assumption that LLM augmentation is consistently superior. Building on empirical findings, we propose Robust Graph Learning via Retrieval-Augmented Contrastive Refinement (RoGRAD) framework. Unlike prior one-shot LLM-as-Enhancer designs, RoGRAD is the first iterative paradigm that leverages Retrieval-Augmented Generation (RAG) to inject retrieval-grounded augmentations by supplying class-consistent, diverse  augmentations and enforcing discriminative representations through iterative graph contrastive learning. It transforms LLM augmentation for graphs from static signal injection into dynamic refinement. Extensive experiments demonstrate RoGRAD's superiority over both conventional GNN- and LLM-enhanced baselines, achieving up to 82.43\% average improvement.
\end{abstract}

%    <concept>
%        <concept_id>10010147.10010257.10010293.10010294</concept_id>
%        <concept_desc>Computing methodologies~Neural networks</concept_desc>
%        <concept_significance>500</concept_significance>
%        </concept>
%  </ccs2012>
% \ccsdesc[500]{Computing methodologies~Neural networks}

% \keywords{Graph Neural Network, Large Language Models, GNN Robustness, LLM-on-graph, Iterative Retrieval-Augmented Generation}

\maketitle

\section{Introduction}\label{Introduction}
Graph Neural Networks (GNNs) have become a cornerstone for analyzing various types of Web graphs with applications ranging from social and traffic networks ~\cite{sharma2024survey,zhang2022improving,yu2018spatio,wang2025multigran}, to recommender systems~\cite{wu2022graph,gao2022graph,huang2021mixgcf}. Although GNNs achieve excellent results on standard benchmark datasets, their performance can degrade substantially when faced with real-world graph data that contain inherent imperfections~\cite{ju2024survey,dai2024comprehensive}. This is because complete graphs in practice suffer from data corruptions and loss, so that labels, structures, and features are often scarce and incomplete, and nodes may be missing or isolated. Such deficiencies, referred to as ‘weak information’~\cite{liu2023learning}, are shown to degrade GNN performance markedly~\cite{you2020handling,taguchi2021graph}.

Beyond solitary studies of such deficiencies, their coexistence is pervasive in practical applications and poses challenges to the reliability and robustness of GNN models. For instance, intact social networks can develop deficiencies when certain social relationships are intentionally hidden by users, and when user profiles or attributes become unavailable due to privacy restrictions or account deactivation.\cite{kossinets2006effects,kim2025accurate,zhou2025common}. Such incompleteness hinders neighborhood aggregation in GNNs and degrades task performance.

Many studies have sought to enhance GNN robustness when graphs are noisy, incomplete, or weakly supervised. Prior works refine graph structures to reduce incompleteness and noise~\cite{jin2020graph,li2022reliable,yuan2024mitigating}, design architectures for robust representations~\cite{zhu2021deep}, or employ training strategies~\cite{feng2019graph,bojchevski2019certifiable} and supervision-efficient learning~\cite{wang2021certified,ding2022meta}. Although these efforts demonstrate progress, their effectiveness under joint degraded conditions is insufficiently investigated.

Recent advances in LLMs endow LLMs with extensive knowledge, proficiency in semantic understanding, and strong reasoning capabilities~\cite{chang2024large,bhargava2022commonsense,hu2023survey}. These properties motivate the integration of LLMs and graphs~\cite{chen2024exploring} either as encoders of node attributes~\cite{khoshraftar2025graphit,yu2023empower,jin2024large}, enhancers providing auxiliary signals~\cite{he2023harnessing,yu2025leveraging}, or reasoners reformulating graph tasks into text~\cite{chai2023graphllm,zhao2023graphtext}.

However, a critical question remains: \textit{Can LLM-enhanced methods truly deliver better supportive performance and stronger robustness than traditional ones under real-world deficiencies?} The comparative merits of LLM-enhanced versus conventional approaches under graph deficiencies remains unexplored. While LLMs bring semantic priors and external knowledge, they incur substantial computational overhead, distinct instability across runs, and hallucinations~\cite{brown2020language,huang2025survey}. Given the trade-offs, the question of whether LLMs are indeed beneficial GNN helpers must be answered.

To address this gap, we empirically benchmark LLM-enhanced and conventional graph-native GNN approaches across a spectrum of graph deficiencies. Centering on the canonical node classification task, we provide the first systematic evaluation of robustness to compound imperfections, uncovering the performance disparities between LLM-enhanced and conventional GNN approaches, offering new insights of incorporating LLMs on graphs.

Our empirical analysis reveals: \textit{Under less severe deficiencies, LLM-enhanced works trail behind non-LLM counterparts that are simpler and more resource-efficient}, we further uncover two key limitations of existing LLM-enhanced frameworks. First, LLM generations suffer from semantic homogeneity rather than providing informative diversity (see Appendix Section~\ref{Issues} for details). This severely constrains GNNs to learn discriminative boundaries, impairing downstream tasks such as node classification and link prediction~\cite{feng2024imo,lu2024mitigating}. Generations across different classes tend to converge on similar phrasing, blurring inter-class distinctions, while samples within the same class are insufficiently coherent, providing heterogeneous signals that fail to consolidate class prototypes. The lack of differentiation weakens the discriminative capacity of encoded representations and reduces augmentation utility.

Second, existing LLM-as-Enhancer frameworks adopt a one-shot paradigm, where augmentations are generated once. This static pipeline fundamentally diminishes GNN performance and robustness, as low-quality or homogeneous generations cannot be refined.
In contrast, we propose the first iterative refinement paradigm for LLM-on-graph that represents a paradigm shift from one-shot enhancement to iterative refinement for LLM-on-graph frameworks.

To address both limitations, we propose \textbf{RoGRAD} (Robust Graph Learning via Retrieval-Augmented Contrastive Refinement), the first iterative refinement framework for LLM-on-graph. RoGRAD leverages a retrieval–diagnosis–revision mechanism to mitigate semantic homogeneity and promote class-consistent diversity at the LLM generation level. Through iterative sample generations, it enriches node features, supplies informative supervisory signals and edges, and expands the limited node set alleviating weak information. In parallel, we further introduce \textbf{R2CL} (\underline{\textbf{C}}ontrastive \underline{\textbf{L}}earning with \underline{\textbf{R}}AG \underline{\textbf{R}}efinement), a novel representation learning paradigm with RAG that enforces semantic alignment and separation in the embedding space through contrastive learning on LLM-perturbed views. Together, these components produce more discriminative and robust node representations from intact data to mitigate performance degradation under compound graph deficiencies.

Our contributions can be summarized as follows:
\begin{itemize}
\item We present the first comprehensive exploration of LLM-on-graph frameworks under varying graph deficiencies and their comparison against non-LLM GNN-based counterparts. Our findings reveal that LLM-enhanced approaches, despite their semantic richness, may underperform simpler non-LLM ones under low-to-modest perturbations, challenging the prevailing assumptions about their superiority.
\item To the best of our knowledge, RoGRAD is the first iterative RAG framework for graph learning tasks that jointly optimizes LLM-generated content and node representations through self-retrieval. It reinforces GNN robustness against common graph deficiencies in real-world practice.
\item At the representation level, we introduce R2CL, a novel contrastive learning framework that leverages iterative retrieval-guided, LLM-refined views to promote intra-class representation consistency and inter-class discriminability.
\item Extensive experiments on benchmark datasets validate the consistent superiority of RoGRAD over selected baselines.
\end{itemize}

\begin{table*}[t]
  \caption{Accuracy (\%) vs. attack intensities. The darkest/lighter shadings highlight the best/second- and third-best performances.}
  \label{empirical_study}
  \centering
  \renewcommand{\arraystretch}{0.9}
  \resizebox{\linewidth}{!}{%
  \begin{tabular}{l*{14}{c}}
    \toprule
    \textbf{Atk. Int.} & \textbf{GCN} & \textbf{ACM-GNN} & \textbf{GRCN} & \textbf{Nbr. Mean} & \textbf{Feat. Prop.} & \textbf{DropMsg} & \textbf{DropEdge} & \textbf{DGI} & \textbf{RS-GNN} & \textbf{TAPE} & \textbf{TA\_E} & \textbf{LLM4NG} & \textbf{MistralEmb} & \textbf{QwenEmb} \\
    \midrule
    0.00          & 87.98 & \cellcolor{gray!25}\textbf{88.54}   & 85.66 & \cellcolor{gray!10}{88.17}     & \cellcolor{gray!10}{88.17}      & 87.65    & 87.61    & 82.88 & 63.59  & 84.49 & 83.36 & 88.06  & 87.87      & 87.69      \\
0.33          & \cellcolor{gray!10}{87.06} & \cellcolor{gray!10}{87.39}   & 86.32 & 86.36     & 86.36      & \cellcolor{gray!25}\textbf{87.43}    & 86.77    & 80.59 & 69.43  & 84.53 & 83.32 & 86.21  & 86.80      & 86.80      \\
0.50          & 83.19 & \cellcolor{gray!25}\textbf{84.81}   & 81.70 & \cellcolor{gray!10}{83.37}     & \cellcolor{gray!10}{83.58}      & 82.23    & 83.68    & 75.17 & 63.44  & 81.74 & 80.59 & 84.07  & 85.27      & 85.56      \\
0.66          & 84.44 & 84.55   & 84.18 & \cellcolor{gray!25}\textbf{85.63}     & \cellcolor{gray!25}\textbf{85.63}      & 84.21    & 83.55    & 74.30 & 78.56  & \cellcolor{gray!10}{85.36} & 83.62 & 84.33  & 86.32      & 85.47      \\
0.83          & 82.26 & \cellcolor{gray!25}\textbf{83.50}   & 82.13 & 82.56     & \cellcolor{gray!10}{82.80}      & 81.83    & 82.33    & 73.01 & 69.69  & 80.49 & 78.58 & \cellcolor{gray!10}{83.49}  & 83.81      & 83.04      \\
0.90          & 72.63 & \cellcolor{gray!25}\textbf{76.51}   & 61.83 & 74.03     & 74.74      & 73.76    & 68.13    & 54.40 & 64.62  & \cellcolor{gray!10}{74.89} & 73.55 & \cellcolor{gray!10}{76.38}  & 80.37      & 79.94      \\
1.00          & \cellcolor{gray!10}{77.11} & \cellcolor{gray!10}{78.38}   & 75.12 & 75.54     & 76.16      & 75.02    & 76.26    & 66.16 & 58.11  & 76.99 & 75.89 & \cellcolor{gray!25}\textbf{80.58}  & 83.92      & 83.50      \\
1.16          & 78.68 & 79.19   & \cellcolor{gray!25}\textbf{81.18} & 79.97     & 80.00      & 78.10    & 77.66    & 64.60 & 71.83  & \cellcolor{gray!10}{80.31} & 78.00 & \cellcolor{gray!10}{80.77}  & 82.26      & 81.66      \\
1.23          & 69.86 & \cellcolor{gray!10}{73.02}   & 61.60 & 68.79     & 69.79      & 69.35    & 63.31    & 53.14 & 63.23  & \cellcolor{gray!10}{71.16} & 69.22 & \cellcolor{gray!25}\textbf{76.13}  & 78.11      & 79.24      \\
1.33          & 75.09 & \cellcolor{gray!10}{77.61}   & 76.21 & \cellcolor{gray!10}{76.27}     & 76.05      & 74.58    & 72.57    & 64.69 & 62.78  & 76.42 & 74.72 & \cellcolor{gray!25}\textbf{80.37}  & 79.84      & 79.86      \\
1.40          & 62.87 & 66.26   & 54.35 & 60.71     & 62.48      & 62.10    & 56.95    & 48.45 & 53.73  & \cellcolor{gray!10}{69.54} & \cellcolor{gray!10}{67.18} & \cellcolor{gray!25}\textbf{73.02}  & 79.00      & 79.06      \\
1.50          & 68.37 & \cellcolor{gray!10}{69.63}   & 66.52 & 62.98     & 64.11      & 65.26    & 63.33    & 56.22 & 50.59  & \cellcolor{gray!10}{71.59} & 67.04 & \cellcolor{gray!25}\textbf{77.19}  & 82.07      & 82.52      \\
1.56          & 61.12 & 64.36   & 58.95 & 63.19     & 64.59      & 61.12    & 56.15    & 45.39 & 60.61  & \cellcolor{gray!10}{70.93} & \cellcolor{gray!10}{67.08} & \cellcolor{gray!25}\textbf{73.57}  & 76.38      & 74.32      \\
1.66          & 70.62 & 71.71   & \cellcolor{gray!10}{73.94} & 72.01     & 71.94      & 69.62    & 67.76    & 55.97 & 62.36  & \cellcolor{gray!10}{73.50} & 70.82 & \cellcolor{gray!25}\textbf{77.50}  & 78.85      & 78.80      \\
1.73          & 60.62 & 62.99   & 54.00 & 56.51     & 58.11      & 59.21    & 52.40    & 46.95 & 53.55  & \cellcolor{gray!10}{66.86} & \cellcolor{gray!10}{65.77} & \cellcolor{gray!25}\textbf{72.18}  & 76.06      & 77.50      \\
1.80          & 48.53 & 47.56   & 42.14 & 42.36     & 43.36      & 46.46    & 39.38    & 36.66 & 39.11  & \cellcolor{gray!10}{59.05} & \cellcolor{gray!10}{59.92} & \cellcolor{gray!25}\textbf{62.19}  & 75.98      & 74.50      \\
1.83          & 66.82 & \cellcolor{gray!10}{69.70}   & 64.37 & 64.52     & 63.87      & 66.59    & 61.04    & 55.78 & 52.89  & \cellcolor{gray!10}{67.11} & 61.41 & \cellcolor{gray!25}\textbf{77.26}  & 77.41      & 78.15      \\
1.90          & 51.58 & 55.06   & 49.09 & 44.36     & 49.14      & 50.66    & 46.42    & 41.36 & 42.10  & \cellcolor{gray!10}{62.51} & \cellcolor{gray!10}{58.15} & \cellcolor{gray!25}\textbf{69.78}  & 79.14      & 77.48      \\
2.06          & 52.25 & 54.93   & 50.90 & 50.17     & 52.22      & 51.53    & 46.00    & 40.53 & 48.99  & \cellcolor{gray!10}{64.16} & \cellcolor{gray!10}{61.32} & \cellcolor{gray!25}\textbf{70.00}  & 73.25      & 74.38      \\
2.13          & 43.29 & 41.63   & 42.99 & 39.02     & 39.74      & 43.06    & 37.22    & 35.75 & 36.35  & \cellcolor{gray!10}{58.44} & \cellcolor{gray!10}{60.12} & \cellcolor{gray!25}\textbf{60.51}  & 75.52      & 74.66      \\
2.16          & 60.22 & 62.29   & \cellcolor{gray!10}{64.22} & 58.06     & 55.89      & 61.26    & 54.30    & 48.30 & 51.04  & \cellcolor{gray!10}{65.37} & 64.00 & \cellcolor{gray!25}\textbf{72.07}  & 75.93      & 75.11      \\
2.23          & 50.89 & 53.31   & 45.41 & 41.93     & 45.39      & 49.78    & 41.16    & 40.22 & 41.95  & \cellcolor{gray!10}{62.90} & \cellcolor{gray!10}{61.78} & \cellcolor{gray!25}\textbf{69.26}  & 74.12      & 74.97      \\
2.30          & 38.15 & 35.85   & 37.21 & 35.49     & 34.37      & 36.29    & 32.66    & 32.52 & 33.83  & \cellcolor{gray!10}{57.22} & \cellcolor{gray!10}{54.81} & \cellcolor{gray!25}\textbf{58.34}  & 76.59      & 75.04      \\
2.46          & 37.50 & 35.23   & 37.49 & 35.62     & 36.00      & 33.82    & 28.69    & 29.53 & 33.26  & \cellcolor{gray!10}{51.54} & \cellcolor{gray!10}{51.01} & \cellcolor{gray!25}\textbf{59.46}  & 66.41      & 69.55      \\
2.56          & 44.40 & 45.85   & 44.42 & 35.67     & 38.93      & 42.77    & 37.48    & 36.72 & 39.28  & \cellcolor{gray!10}{59.01} & \cellcolor{gray!10}{56.96} & \cellcolor{gray!25}\textbf{64.27}  & 71.26      & 73.26      \\
2.63          & 34.00 & 33.66   & 35.97 & 29.58     & 30.09      & 32.59    & 32.39    & 31.11 & 32.22  & \cellcolor{gray!10}{55.63} & \cellcolor{gray!10}{55.11} & \cellcolor{gray!25}\textbf{59.33}  & 72.57      & 73.51      \\
2.70          & 33.34 & 23.70   & 29.26 & 35.51     & 35.51      & 27.04    & 29.63    & 29.63 & 28.89  & \cellcolor{gray!10}{50.18} & \cellcolor{gray!25}\textbf{51.85} & \cellcolor{gray!10}{49.26}  & 74.81      & 71.11      \\
2.96          & 33.23 & 30.52   & 32.25 & 28.05     & 28.48      & 30.25    & 28.62    & 29.51 & 30.37  & \cellcolor{gray!10}{49.96} & \cellcolor{gray!10}{47.41} & \cellcolor{gray!25}\textbf{54.30}  & 67.58      & 68.94      \\
3.03          & 29.63 & 22.22   & 29.26 & 29.39     & 29.79      & 27.78    & 29.63    & 28.52 & 28.15  & \cellcolor{gray!10}{51.66} & \cellcolor{gray!25}\textbf{53.33} & \cellcolor{gray!10}{50.00}  & 74.81      & 71.85      \\
3.36          & 27.41 & 19.63   & 26.30 & 29.39     & 29.39      & 23.70    & 27.41    & 26.67 & 24.81  & \cellcolor{gray!10}{40.37} & \cellcolor{gray!10}{40.74} & \cellcolor{gray!25}\textbf{44.44}  & 62.96      & 67.41 \\
    \bottomrule
  \end{tabular}}
\end{table*}

\section{Related Work} \label{Related Work}
\subsection{Graph Learning with Deficiencies}
Over the years, various methods have been proposed to enhance GNN performance under weak information. Relation-aware models~\cite{schlichtkrull2018modeling,vashishth2019composition} extend GNN architectures by modeling relations to compensate for incomplete and noisy structures. ACM-GNN~\cite{luan2022revisiting} proposes an adaptive channel mixing strategy, which relieves homophilous edge sparsity. Structure refinement models such as GRCN~\cite{yu2020graph,franceschi2019learning} mitigate fixed structures by dynamically revising graph topologies, benefiting scenarios characterized by sparse and spurious edges. While perturbation techniques such as DropEdge or feature masking~\cite{rong2019dropedge,papp2021dropgnn,tan2022supervised} are effective in improving generalization and solving overfitting, semi-supervised and few-shot methods~\cite{kim2019edge,ding2022meta,lee2022grafn} alleviate the problem of limited labels. DropMessage~\cite{fang2023dropmessage} perturbs node-to-node communication for boosting robustness in message-passing. Likewise, feature propagation~\cite{rossi2022unreasonable} and imputation (e.g., Neighbor Mean) achieve strong performance with missing features. RSGNN~\cite{dai2022towards} tackles both edge and supervision deficiency by down-weighting noisy edges and densifying the graph to exploit scarce labels. Besides, unsupervised methods such as DGI~\cite{velivckovic2018deep} cope with label deficiency by learning without labels. However, these approaches address specific deficiencies, and their effectiveness under compound deficiencies remains underexplored.
  
\subsection{LLM-on-graph}
Emerging paradigms have proliferated through the integration of LLMs and graphs. LLM-as-Encoder paradigm leverages pretrained LLMs to transform textual attributes into semantically richer and higher-dimensional embeddings, compared to shallow embeddings (e.g., bag-of-words)~\cite{wu2025llms,khoshraftar2025graphit,yu2023empower}, whereas LLM-as-Enhancer augments GNNs with LLM-generated contents. TAPE~\cite{he2023harnessing} distills textual explanations from GPT into compact embeddings that improve node classification, while LLM4NG~\cite{yu2025leveraging} generates labeled nodes for augmenting graphs under label-limited few-shot settings. Similarly, Chen et al.~\cite{chen2024exploring} show that textual features derived from frozen LLMs enrich node representations under various training paradigms. LLM-only reasoning frameworks, including LLM-as-Predictor, bypass GNNs entirely and delegate reasoning to LLMs. It's important to note that, as LLM-only reasoning frameworks such as GraphText~\cite{zhao2023graphtext} and GraphLLM~\cite{chai2023graphllm} operate independently of graph encoders, their performance is not directly comparable to that of GNN-based approaches, lying outside our scope.

\subsection{GNN Robustness with LLMs}
To date, the potential of LLM-on-graph and investigations on GNN robustness with LLMs are still underexplored. The only related work, Zhang et al.~\cite{zhang2025can}, mainly examines whether LLM-as-Encoder improves adversarial robustness. However, they focus on adversarial robustness where sophisticated attackers launch crafted perturbations, while our work emphasizes graph defect robustness commonly encountered in real-world applications, towards more practical implications. Besides, their perturbation scope is narrowly limited to structural adversarial attacks, with feature, node outage, and supervision attacks unstudied. Most importantly, LLMs cannot be the only augmentation solutions, and they are not systematically benchmarked against conventional GNN-based approaches in ~\cite{zhang2025can}.

\section{GNN Robustness against Graph Deficiencies}
To understand how graph deficiencies affect GNN performance, we evaluate a diverse set of baselines including GNN-based and recent LLM-enhanced techniques against various deficiencies.

\subsection{Preliminaries}
\subsubsection{Node Classification with Text-attributed Graphs}
Let $\mathcal{G} = (\mathcal{V}, \mathcal{E}, \mathcal{T})$ denote a text-attributed graph, where $\mathcal{V} = \{v_1, \ldots, v_n\}$ and $\mathcal{E} \subseteq \mathcal{V} \times \mathcal{V}$ are the sets of nodes and edges, $\mathcal{T} = \{t_1, \ldots, t_n\}$ denotes the set of node attributes. Each node $v_i \in \mathcal{V}$ is coupled with corresponding texts $t_i \in \mathcal{T}$, such as descriptions. To process such textual information, a function $\phi: \mathcal{T} \rightarrow \mathbb{R}^d$ is employed to encode each $t_i$ into a representation $\mathbf{x}_i = \phi(t_i)$, yielding a node feature matrix $\mathbf{X} \in \mathbb{R}^{n \times d}$ stacking all embeddings. Structural information is captured by an adjacency matrix $\mathbf{A} \in \{0,1\}^{n \times n}$, where $A_{ij} = 1$ if an edge exists. In the semi-supervised setting, a node subset $\mathcal{V}_L \subset \mathcal{V}$ is associated with ground-truth labels $\{y_i \mid v_i \in \mathcal{V}_L\}$, where each $y_i \in \mathcal{Y}$ and $|\mathcal{Y}| = C$, with $C$ denoting the total number of classes. The remaining nodes $\mathcal{V}_U = \mathcal{V} \setminus \mathcal{V}_L$ are unlabeled, and the final goal is to predict their labels by learning a mapping $f: (\mathbf{A}, \mathbf{X}) \rightarrow \mathcal{Y}^n$.

\subsubsection{GNN robustness against graph deficiencies}
In real-world scenarios, the observed graph $\mathcal{G}_{\text{obs}} = (\mathcal{V}, \mathcal{E}_{\text{obs}})$ and node features $\mathbf{X}_{\text{obs}} \in \mathbb{R}^{n \times d}$ are rarely ideal. First, supervision is commonly constrained to a small subset $\mathcal{V}_L \subset \mathcal{V}$, having known labels $\{y_i\}_{v_i \in \mathcal{V}_L}$, $|\mathcal{V}_L| \ll |\mathcal{V}|$. The structural observation $\mathcal{E}_{\text{obs}} \subset \mathcal{E}_{\text{true}}$ leads to incomplete adjacency $\mathbf{A}_{\text{obs}} \subset \mathbf{A}_{\text{true}}$. Feature vectors can be partially missing with $\exists j: x_{ij} = \varnothing$. Finally, available nodes may be insufficient for learning meaningful representations when $|\mathcal{V}|$ is small or sparsely connected. These deficiencies imply that $(\mathbf{A}_{\text{obs}}, \mathbf{X}_{\text{obs}})$ deviates from the intact graph $(\mathbf{A}_{\text{true}}, \mathbf{X}_{\text{true}})$. GNNs are therefore required to satisfy $f(\mathbf{A}_{\text{obs}}, \mathbf{X}_{\text{obs}}) \approx f(\mathbf{A}_{\text{true}}, \mathbf{X}_{\text{true}})$ to establish robustness for reliable node representations and classification.

\begin{figure}[b]
  \setlength{\abovecaptionskip}{7.5pt}
  \centering
  \includegraphics[width=\linewidth,height=4.2cm]{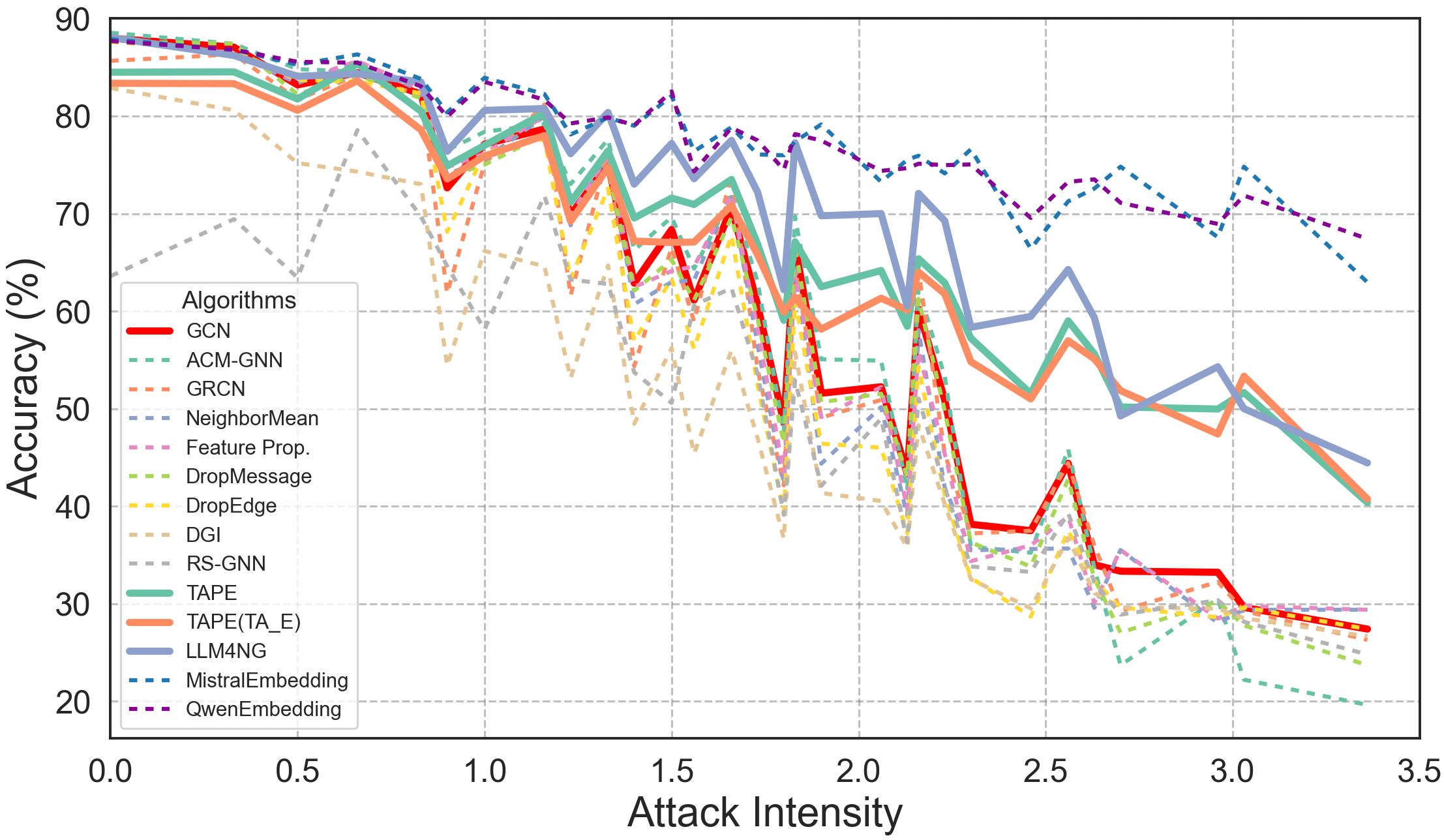}
  \caption{Accuracy under increasing attack intensities.}
  \label{Exploration_fig}
\end{figure}

\subsubsection{Baselines}
\sloppy
We categorize baselines introduced in Section~\ref{Related Work} according to graph deficiencies they directly or potentially mitigate. 

\textbf{Structural Deficiency:} RSGNN~\cite{dai2022towards}, GRCN~\cite{yu2020graph} and ACM-GNN~\cite{luan2022revisiting} enhance graph structure robustness against missing or noisy edges.
\textbf{Supervision Deficiency:} DGI~\cite{velivckovic2018deep},
RSGNN~\cite{dai2022towards} and LLM4NG~\cite{yu2025leveraging} cope with the scarcity of labeled data.
\textbf{Feature Deficiency:} Neighbor Mean, Feature Propagation~\cite{rossi2022unreasonable}, TAPE~\cite{he2023harnessing} and TA\_E~\cite{he2023harnessing} improve robustness to missing or corrupted node features.
\textbf{Node Deficiency:} DropEdge~\cite{rong2019dropedge} and DropMessage~\cite{fang2023dropmessage} help to learn robust representations with part of the nodes missing, since over-fitting weakens the generalization on small graphs.

We also adopt Mistral-7B~\cite{jiang2023mistral7b} and Qwen-3B~\cite{bai2023qwen} following the LLM-as-Encoder paradigm as baselines. All of the above baselines in the empirical study are built on top of a standard GCN backbone.

\subsubsection{Empirical Study Settings}\label{experiment_settings}
We conduct our empirical analysis on the Cora dataset~\cite{yang2016revisiting}.
To ensure fair and consistent comparisons, all base GNN models (GCN~\cite{kipf2016semi}, GAT~\cite{velivckovic2017graph}, and GraphSAGE~\cite{hamilton2017inductive}) are implemented with identical configurations, see Appendix Section~\ref{BaseGNNConf} for details. For all LLM-as-Enhancer baselines, Sentence-BERT~\cite{reimers2019sentence} is employed to encode LLM-generated contents. 

\begin{figure}[t]
  \centering
  \includegraphics[width=7.5cm,height=6cm]{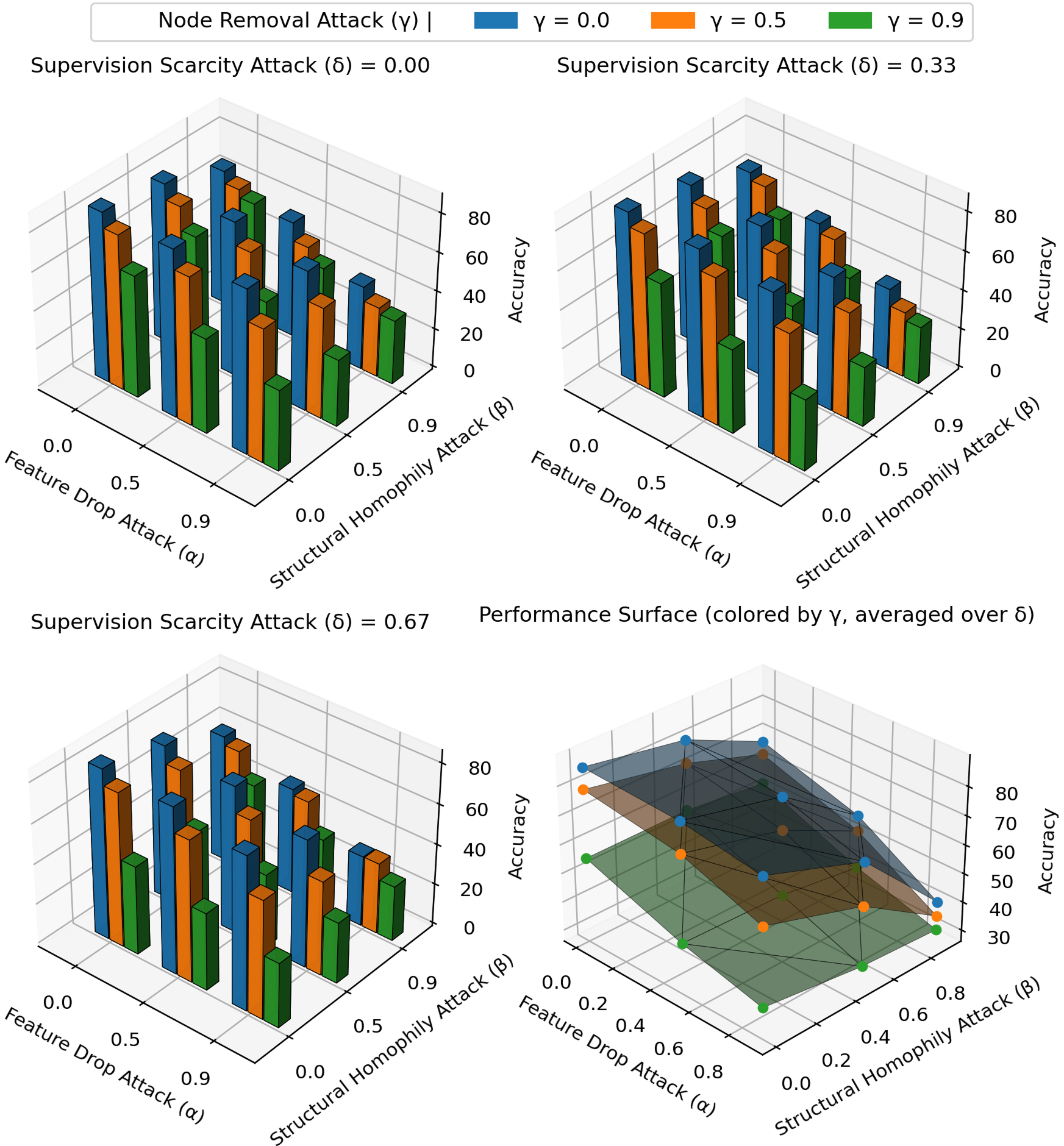}
  \caption{GCN Performance under compound deficiencies.}
  \setlength{\abovecaptionskip}{1pt}
  \label{3D}
\end{figure}

\subsection{Empirical Analysis}
Four types of deficiencies are injected on Cora before training the downstream GNN classifiers. We report the perturbation space: Structural Homophily Attack (SHA) with edge reduction ratio $\in \{0,0.5,0.9\}$, Supervision Scarcity Attack (SSA) with labeled training ratio $\in \{0.6,0.4,0.2\}$, Feature Drop Attack (FDA) with feature drop ratio $\in \{0,0.5,0.9\}$, and Node Removal Attack (NRA) with node drop ratio $\in \{0,0.5,0.9\}$. Validation and test sets are fixed at 20\%.

Our analysis addresses: 
(1) \textit{How do compound deficiencies impact baselines} (Obs. 1-2)? 
(2) \textit{Does LLM-as-Enhancer provides performance and robustness gains worth their complexity} (Obs. 3-5)?
(3) \textit{Do baselines exhibit uniform or deficiency-specific sensitivity} (Obs. 6)?

\underline{\textbf{Observation 1.}} \textbf{Compound deficiencies induce amplified performance degradation.}
Table~\ref{empirical_study}, which aggregates the exhaustive results by summing intensities of individual attacks from full tables exemplified by Appendix Table~\ref{full_table}, shows that accuracy consistently decreases as attack intensity increases. This cumulative impact exceeds what any single deficiency would cause in isolation. Taking vanilla GCN for illustration, the decline in both its performance visualization Figure~\ref{3D} and Table~\ref{empirical_study} further becomes steeper and non-linear when multiple deficiencies co-occur at high levels, revealing a compounding effect that accelerates performance degradation. This necessitates jointly analyzing graph deficiencies rather than in isolation, a perspective rarely studied before.

\underline{\textbf{Observation 2.}} \textbf{Classic GNN is strong baseline.} Classic message-passing GNN (GCN) exhibits strong performance against almost all non-LLM baselines under compound graph deficiencies, which aligns with the findings of Luo et al.~\cite{luo2024classic}. Our results extend this evidence to robustness studies by demonstrating classic GCN's competitiveness when evaluated with a large suite of attacks.

\underline{\textbf{Observation 3.}} \textbf{LLM-as-Encoder exhibits superior robustness by providing richer high-dimensional embeddings.} As shown in Table~\ref{empirical_study}, Mistral and Qwen Embeddings sustain higher accuracy, showing only gradual declines under strong attacks. The exhibited robustness can be attributed to the richer and higher-dimensional embeddings that LLMs' hidden layers provide. As LLM-based encoders use fundamentally different pretrained embeddings and are not included in subsequent comparisons for fairness.

\underline{\textbf{Observation 4.}} \textbf{LLM augmentations fall behind simple GNN counterparts under modest deficiencies.} Table~\ref{empirical_study} and Figure~\ref{Exploration_fig} show that across low-to-moderate attack regimes, both LLM-as-Encoder and LLM-as-Enhancer paradigms perform worse-to-comparable accuracy than much simpler GNN-based counterparts. This observation demonstrates that the integration of LLMs does not consistently yield better performance and can be even worse in many situations. Given the trade-offs mentioned in Section~\ref{Introduction}, especially compared with extreme naive approaches such as Feature Propagation, Neighbor Mean, and DropMessage, LLMs are not always efficient GNN helpers against graph deficiencies.

\begin{table}[b]
\centering
\caption{Robustness evaluation. The reported values are averaged over the four single-type attacks. "Clean Acc" denotes the accuracy (\%) without attack, and "AUC Norm" is the normalized area under the accuracy–attack curve.}
\label{experimental_robustness}
\renewcommand{\arraystretch}{0.8}
\resizebox{\linewidth}{!}{%
\begin{tabular}{lcccc}
\toprule
Algorithm & Clean Acc & Worst Acc & Avg Acc & Norm-AUC \\
\midrule
GCN              & 87.98 & 75.585 & 82.575 & 0.833 \\
DropEdge         & 87.61 & 71.988 & 81.351 & 0.825 \\
DropMessage      & 87.65 & 76.370 & 82.517 & 0.831 \\
Feat. Prop.    & \cellcolor{gray!10}88.17 & 77.462 & 83.303 & 0.838 \\
NeighborMean     & \cellcolor{gray!10}88.17 & 76.930 & 83.072 & 0.836 \\
ACM-GNN          & \cellcolor{gray!25}\textbf{88.54} & \cellcolor{gray!10}78.518 & \cellcolor{gray!10}84.170 & \cellcolor{gray!10}0.847 \\
DGI              & 82.88 & 59.375 & 72.928 & 0.744 \\
RS-GNN           & 63.59 & 61.118 & 65.543 & 0.654 \\
GRCN             & 85.66 & 67.415 & 78.643 & 0.802 \\
MistralEmb.       & 87.87 & \cellcolor{gray!25}\textbf{81.718} & \cellcolor{gray!25}\textbf{85.125} & \cellcolor{gray!25}\textbf{0.854} \\
QwenEmb.          & 87.69 & \cellcolor{gray!10}81.238 & \cellcolor{gray!10}84.960 & \cellcolor{gray!10}0.853 \\
TAPE             & 84.49 & 77.155 & 81.480 & 0.819 \\
TAPE (TA\_E)     & 83.36 & 75.472 & 80.232 & 0.807 \\
LLM4NG           & 88.06 & 78.285 & 83.678 & 0.842 \\
\bottomrule
\end{tabular}}
\end{table}

\underline{\textbf{Observation 5.}} \textbf{LLM-as-Enhancer (e.g., LLM4NG, TAPE, TA\_E) shows no clear robustness advantage over conventional GNN baselines.} While LLM Encoders achieve the highest robustness metrics in Table~\ref{experimental_robustness}, they operate on a fundamentally different representational basis. LLM-as-Enhancer baselines exhibit markedly weaker robustness. Their Worst Acc, Avg Acc, and Norm-AUC scores lag behind or show no improvement over a substantial set of conventional GNN approaches. This finding points out that despite their higher architectural complexity, current LLM-as-Enhancer frameworks fail to deliver tangible robustness gains.

\underline{\textbf{Observation 6.}} \textbf{Different baselines exhibit distinct sensitivity profiles to deficiency types.}
On Cora, the examined baselines exhibit distinct vulnerability patterns. For instance, ACM-GNN, DropEdge, and Neighbor Mean are highly sensitive to NRA and SHA, showing steep performance degradation, but are less affected by FDA. In contrast, LLM4NG and TAPE are relatively more stable under FDA and SHA but show moderate sensitivity to high-intensity NRA. RS-GNN is relatively robust to FDA but performs poorly overall. Therefore, robustness is highly algorithm-dependent, and different baselines present disparate sensitivity.

In summary, current LLM-as-Enhancer frameworks remain complex yet underperform under low-to-moderate perturbations and show weaker robustness by the Norm-AUC metric. Algorithms also differ in sensitivity to deficiency composition, exposing a robustness gap that calls for more resilient approaches.

\begin{figure*}[t]
  \centering
  \includegraphics[width=\linewidth,height=6.5cm]{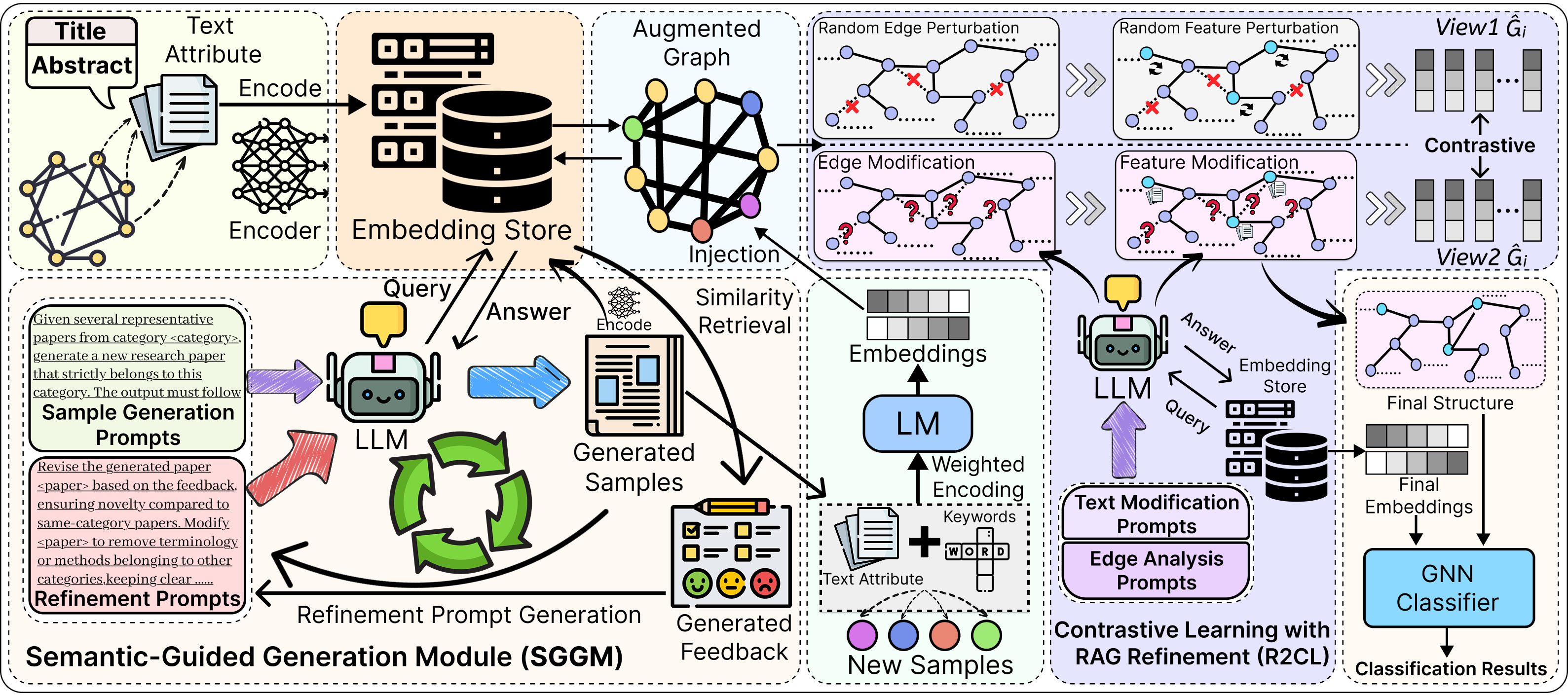}
  \caption{Overall architecture of RoGRAD. RoGRAD establishes the first iterative RAG+GCL paradigm for LLM-on-graph, replacing static one-shot augmentation with dynamic multi-round refinement.}
  \label{system_pipeline}
\end{figure*}

\section{Methodology} \label{Methodology}
\subsection{System Overview}

Our analysis indicates that LLM-as-Enhancer frameworks fail to deliver better robustness, and occasionally underperform simpler GNN-based augmentation. A key reason lies in the quality of the LLM-generated contents \(\hat{T}=\{\hat{t}_i\}\) derived from the original attributes \(T=\{t_i\}\). First, \(\hat{t}_i\) tend to be semantically incoherent within the same class \(\mathcal{C}_k\) and homogeneous across different classes, due to the lack of semantic guidance. 
After being encoded and propagated in GNNs, the learned representations \(\mathbf{z}_i\) exhibit high intra-class variance $\sigma^2_{\text{intra}}(k)$ and small inter-class margins $m_{\text{inter}}(k,k')$:

\begin{equation}
\sigma^2_{\text{intra}}=\frac{1}{|\mathcal{C}_k|}\sum_{v_i\in\mathcal{C}_k}\|\mathbf{z}_i-\bar{\mathbf{z}}_k\|_2^2\uparrow,\;
m_{\text{inter}}(k,k')=\|\bar{\mathbf{z}}_k-\bar{\mathbf{z}}_{k'}\|_2\downarrow.
\end{equation}

Second, the adoption of one-shot generation produces each \(\hat{t}_i=\mathcal{G}_\theta(t_i)\) with a LLM \(\mathcal{G}_\theta\) in a single pass without refinement. As a result, low-quality or noisy augmentations persist.

To solve such issues, RoGRAD establishes a retrieval-augmented learning pipeline over the text-attributed graph \(G=(\mathcal{V},\mathcal{E},\mathcal{T})\) to enhance graph robustness by improving structural connectivity and semantic discriminability before downstream classification. The overall architecture of RoGRAD is shown in Figure~\ref{system_pipeline}, which employs: 
(i) a \textit{Semantic-Guided Generation Module (SGGM)} that iteratively synthesizes and refines augmentation samples; 
(ii) a \textit{Graph Enrichment Stage} that injects generated samples into original \(G\) to expand node coverage and features, introduce new high-confidence edges, and supplement pseudo-label supervision; 
and (iii) a \textit{Contrastive Learning with RAG Refinement (R2CL)} module that regularizes node representations on randomly and LLM-perturbed contrastive views along both structure and embedding dimensions.

Concretely, SGGM leverages \(\mathcal{G}_\theta\) to produce and iteratively refines initial sample drafts \(\hat{t}_i^{(0)}\) from \(t_i \in \mathcal{V}\), where it retrieves top-\(k\) same-class neighbors from the embedding store 
\(\mathcal{M}=\{\mathbf{e}_i=\phi(t_i)\mid v_i\in V\}\) as semantic groundings. Synthetic samples are then encoded by a Sentence-BERT encoder \(\psi(\cdot)\), 
and a feedback \(\mathcal{F}(\hat{t}_i^{(r)})\) summarizes similarity-based diagnoses derived from comparisons between \(\hat{\mathbf{x}}_i^{(r)}\) and existing nodes. Optimized samples are later merged into the initial graph.
Finally, R2CL constructs LLM-perturbed views $\widetilde{\mathcal{G}}^{(1)}$ and $\widetilde{\mathcal{G}}^{(2)}$ to obtain representations that enforce intra-class alignment and inter-class separation.

\subsection{Semantic-Guided Generation}

SGGM refines the text $t_i$ of augmentation samples $\hat{v}_i \in \hat{V}$ into a high-quality augmentation \(\hat{t}_i\) through a retrieval–diagnosis–revision pipeline driven by LLM \(\mathcal{G}_\theta\), so that generations can be optimized at least once, compared with one-shot generation. Original attributes are encoded as \(\mathbf{e}_j=\phi(t_j)\) and stored in an embedding memory \(\mathcal{M}=\{\mathbf{e}_j\mid v_j\in V\}\), and the initial \(\hat{t}_i^{(0)}\) generated by prompts in Figure~\ref{initial_prompt} is encoded as \(\psi(\hat{t}_i^{(0)})\). SGGM retrieves \(k\) semantically close same-class exemplars \(\mathcal{R}_i^{(r)}\) at round \(r\) as grounding context:
\begin{equation}
\begin{aligned}
\mathcal{R}_i^{(r)} = \operatorname{TopK}_{j\in\mathcal{C}(i)}^{\pi} \mathrm{sim}(\hat{\mathbf{x}}_i^{(r)},\mathbf{e}_j) 
= \operatorname{TopK}\!\left(\Big\langle 
   \tfrac{\hat{\mathbf{x}}_i^{(r)}}{{\|\hat{\mathbf{x}}_i^{(r)}\|_2}},\;
   \tfrac{\mathbf{e}_j}{{\|\mathbf{e}_j\|_2}}
\Big\rangle\right).
\end{aligned}
\end{equation}

\begin{figure}[t]
\centering
\includegraphics[width=\linewidth]{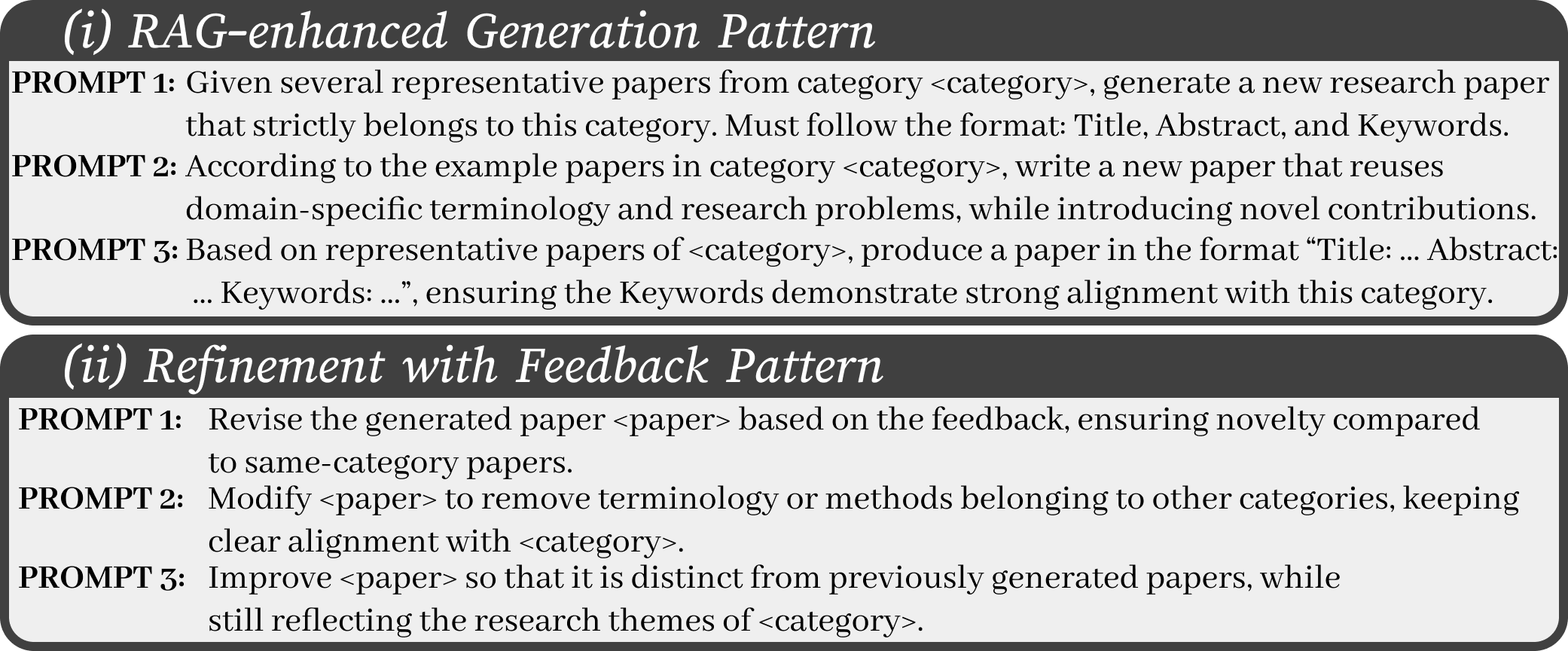}
\caption{Prompts for semantic-guided generation.}
\label{SEG_prompt}
\end{figure}

These exemplars are embedded into \textit{RAG-enhanced generation prompts} that instruct \(\mathcal{G}_\theta\) to: (1) analyze exemplars to extract category-specific terminology, methodologies, and topics; (2) generate a new text using representative terms while addressing a typical same-category problem; and (3) output the text in a strict Title–Abstract–Keywords format, illustrated by Figure~\ref{SEG_prompt} (i). The additional keywords are incorporated into the embedding with a fusion weight $\lambda$ to enhance intra-class alignment. After generation, the new draft \(\hat{t}_i^{(r+1)}\) is then encoded as \(\hat{\mathbf{x}}_i^{(r+1)}=\psi(\hat{t}_i^{(r+1)})\) with main content and keywords. 
Its similarity \(\mathbf{s}_i^{(r+1)}\) against all existing nodes becomes \([\mathrm{sim}(\hat{\mathbf{x}}_i^{(r+1)},\mathbf{x}_j)]_{j\in V}\),
based on which, four diagnostic metrics: redundancy \(r_i^{(r+1)}\), class alignment \(a_i^{(r+1)}\), off-category drift \(o_i^{(r+1)}\), and duplication \(d_i^{(r+1)}\) are computed:
\begin{equation}
\begin{aligned}
r_i^{(r+1)} &= \max_{\scriptscriptstyle j\in\mathcal{C}(i)}
  \mathrm{sim}(\hat{\mathbf{x}}_i^{(r+1)},\mathbf{x}_j),\;
a_i^{(r+1)} = \sum_{\scriptscriptstyle j\in\operatorname{TopK}_{\scriptscriptstyle\mathcal{C}(i)}} 
\tfrac{\mathrm{sim}(\hat{\mathbf{x}}_i^{(r+1)},\mathbf{x}_j)}{k},\\
o_i^{(r+1)} &= \max_{\scriptscriptstyle j\notin\mathcal{C}(i)}
  \mathrm{sim}(\hat{\mathbf{x}}_i^{(r+1)},\mathbf{x}_j),\;
d_i^{(r+1)} = \max_{\scriptscriptstyle \hat{v}_j\in\hat{V}_{\scriptscriptstyle\text{prev}}}
  \mathrm{sim}(\hat{\mathbf{x}}_i^{(r+1)},\hat{\mathbf{x}}_j).
\end{aligned}
\end{equation}

A critique process \(\mathcal{F}\) then converts the diagnostic scores \((r_i^{(r+1)},\,a_i^{(r+1)},\,o_i^{(r+1)},\,d_i^{(r+1)})\)
into targeted refinement instructions. 
Specifically, if \(r_i^{(r+1)}\) or \(d_i^{(r+1)}\) are high, the instruction emphasizes introducing novel elements; 
if \(a_i^{(r+1)}\) is low, it calls for reinforcing category-specific terms; 
and if \(o_i^{(r+1)}\) is high, it instructs to remove off-category terminology. 
The \textit{feedback-based refinement prompts} in Figure~\ref{SEG_prompt} (ii) guide \(\mathcal{G}_\theta\) to rewrite drafts with greater diversity compared to same-class exemplars, stricter category alignment, and minimal overlap with previously generated texts.  

Finally, the original text \(t_i\), the retrieved exemplars \(\mathcal{R}_i^{(r+1)}\), and the feedback \(\mathcal{F}(\hat{t}_i^{(r+1)})\) can be fed to \(\mathcal{G}_\theta\) to produce the next draft:
\begin{equation}
\hat{t}_i^{(r+2)}=\mathcal{G}_\theta\big(t_i,\mathcal{R}_i^{(r+1)},\mathcal{F}(\hat{t}_i^{(r+1)})\big).
\end{equation}
This refinement stops when no violations remain or when the maximum round \(R\) is reached. 
The final \(\hat{T}=\{\hat{t}_i^{(R)}\}\) with embeddings \(\hat{\mathbf{x}}_i=\psi(\hat{t}_i^{(R)})\) is then utilized to enrich the initial graph.

\subsection{Graph Enrichment Against Deficiencies}\label{enrichment}
Refined samples are incorporated to compensate for deficiencies with encoded features $\hat{\mathbf{x}}_i$. We assemble them as $\hat{X}=\big[\hat{\mathbf{x}}_1;\ldots;\hat{\mathbf{x}}_M\big]\in\mathbb{R}^{M\times d}$ and introduce the corresponding nodes $\hat{V}=\{\hat{v}_1,\ldots,\hat{v}_M\}$ into the original graph $\mathcal{G=(V,E,X)}$, yielding an enriched graph $\widetilde{\mathcal{G}} = (\widetilde{\mathcal{V}}, \widetilde{\mathcal{E}},\widetilde{\mathcal{X}})$ with $\widetilde{\mathcal{V}}=V\cup\hat{V}$ and $\widetilde{\mathcal{X}}=[X;\hat{X}]\in\mathbb{R}^{(N+M)\times d}$. 

For structural enrichment, each $\hat{v}_i$ is connected to semantically similar original nodes according to feature similarity $\mathrm{sim}(\hat{\mathbf{x}}_i,\mathbf{x}_j)=\hat{\mathbf{x}}_i^\top\mathbf{x}_j$. The threshold $\tau$-based neighborhood is defined as:
\begin{equation}
\begin{aligned}
&\mathcal{N}_\tau(\hat{v}_i) = \{\,v_j\in V \mid \mathrm{sim}(\hat{\mathbf{x}}_i,\mathbf{x}_j)>\tau\,\},\\
&\mathcal{E}\;\;
\leftarrow \mathcal{E}\cup\{\,(\hat{v}_i,v_j)\mid v_j\in\mathcal{N}_\tau(\hat{v}_i)\,\}.
\end{aligned}
\end{equation} 

For supervision enrichment, each $\hat{v}_i$ is assigned a hard pseudo-label $\hat{y}_i\in\{1,\dots,C\}$ aligned with its generation category, and the label set is updated as $\mathcal{Y}=[Y_{\text{orig}};\hat{Y}]$. To ensure their participation in optimization, we extend the training mask as $\mathcal{\mathbf{m}}^{\text{train}}=[\mathbf{m}^{\text{train}};\mathbf{1}_M]$ while keeping $\mathcal{\mathbf{m}}^{\text{val}}=[\mathbf{m}^{\text{val}};\mathbf{0}_M]$ and $\mathcal{\mathbf{m}}^{\text{te}}=[\mathbf{m}^{\text{te}};\mathbf{0}_M]$, where \(M\) denotes the number of generated nodes newly inserted.

Consequently, the graph evolves from $G$ to $\mathcal{G}$ with $|\mathcal{V}|=N+M$ and $|\mathcal{E}|=|E|+\sum_i|\mathcal{N}_\tau(\hat{v}_i)|$, providing denser connectivity and richer supervisory signals with additional node embeddings. This enrichment effectively alleviates weak graph information to yield a structurally and semantically informative augmented graph.

\begin{figure}[b]
  \centering
  \includegraphics[width=\linewidth,height=3cm]{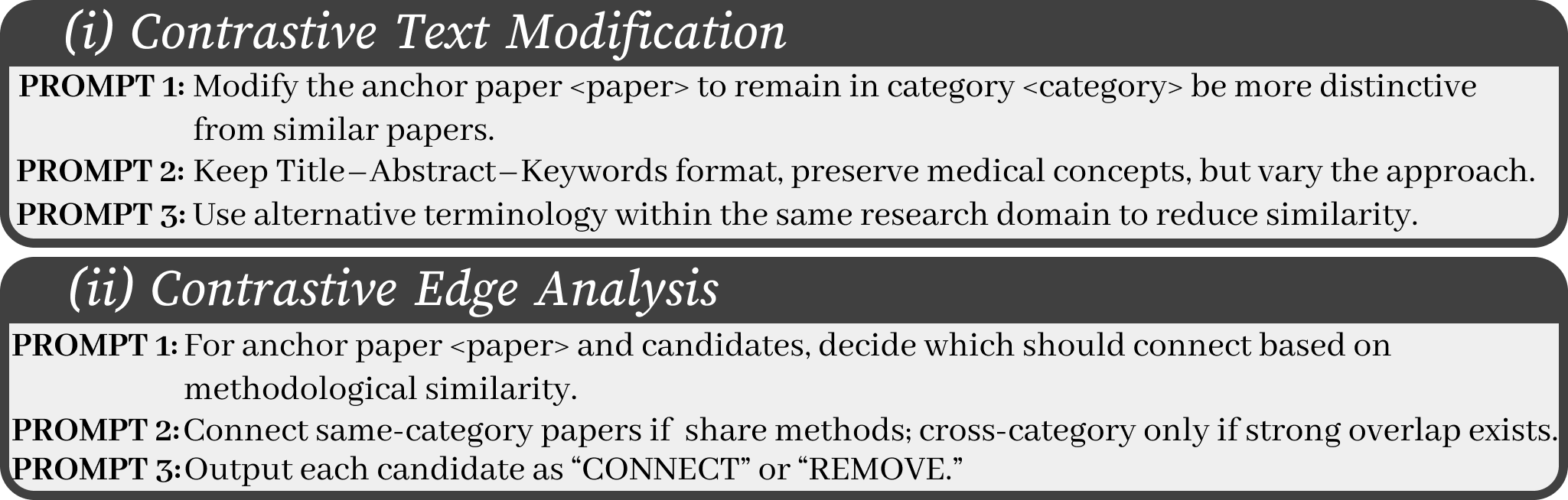}
  \caption{Prompts for retrieval-refined contrastive learning.}
  \label{Contrastive_learning_prompt}
\end{figure}

\begin{table*}[t]
\caption{Overall performance: (a) Improvement over Top 4 baselines on non-deficient graphs, (b) Performance under deficiencies.}
\centering
% ------- left table -------
\begin{minipage}[t]{\linewidth}
\centering
\label{Main_Table}
\renewcommand{\arraystretch}{1.2}
\resizebox{\linewidth}{!}{%
\begin{tabular}{l l|ccccccccccccc|c}
\hline
\textbf{Dataset} & \textbf{Arch.} 
& \textbf{Base GNN}
& \textbf{ACM-GNN}
& \textbf{GRCN}
& \textbf{Nbr. Mean} 
& \textbf{Feat. Prop.}
& \textbf{DropMsg}
& \textbf{DropEdge}   
& \textbf{DGI}
& \textbf{RS-GNN}
& \textbf{TAPE} 
& \textbf{TA\_E} 
& \textbf{LLM4NG} 
& \textbf{Ours} 
& \textbf{Improv.}\\
\hline
\multirow{3}{*}{Cora} 
& GCN        & 87.98 & 88.54 & 85.66 & 88.17 & 88.17 & 87.65 & 87.61 & 82.88 & 63.59 & 84.49 & 83.36 & 88.06 & 89.39 & \textbf{1.31\%} \\
& GAT        & 88.43 & 87.99 & N/A & 87.16 & 87.16 & 88.28 & 86.25 & 83.85 & N/A & 83.86 & 82.74 & 88.47 & 90.17 & \textbf{2.13\%}\\
& SAGE  & 88.84 & 88.91 & 83.62 & 88.25 & 88.25 & 89.06 & 89.24 & 70.50 & 72.09 & 83.90 & 82.59 & 87.06 & 90.39 & \textbf{1.55\%}\\
\hline
\multirow{3}{*}{pubMed} 
& GCN        & 86.36 & 85.45 & 65.72 & 86.10 & 86.09 & 86.21 & 87.60 & 80.97 & 72.97 & 82.14 & 81.43 & 88.25 & 96.29 & \textbf{10.54\%}\\
& GAT        & 87.19 & 81.44 & N/A & 85.69 & 85.73 & 86.35 & 84.57 & 80.73 & N/A & 83.13 & 83.48 & 88.85 & 95.94 & \textbf{10.24\%}\\
& SAGE  & 88.06 & 87.98 & 85.83 & 88.68 & 88.59 & 86.41 & 86.41 & 73.01 & 80.91 & 78.46 & 75.80 & 90.48 & 97.04 & \textbf{9.09\%}\\
\hline
\multirow{3}{*}{Arxiv} 
& GCN        & 72.73 & 75.77 & 66.74 & 77.39 & 80.14 & 66.04 & 71.73 & 63.61 & 46.73 & 83.04 & 82.61 & 82.09 & 86.79 & \textbf{5.88\%}\\
& GAT        & 72.45 & 46.32 & N/A & 80.48 & 80.24 & 70.07 & 42.66 & 64.37 & N/A & 83.30 & 82.57 & 84.51 & 88.79 & \textbf{7.34\%}\\
& SAGE  & 68.51 & 79.09 & 63.33 & 79.48 & 79.71 & 68.03 & 66.70 & 54.06 & 68.77 & 85.22 & 83.52 & 84.99 & 92.16 & \textbf{10.56\%}\\
\hline
\end{tabular}}
\end{minipage}
\hfill

% ------- right table -------
\begin{minipage}[t]{\linewidth}
\centering
\renewcommand{\arraystretch}{1.0}
\resizebox{\linewidth}{!}{%
\begin{tabular}{l|ccc|ccc|ccc|ccc|ccc|ccc|ccc}
\hline
\textbf{Atk. Int.} 
& \multicolumn{3}{c|}{\textbf{ACM-GNN}} 
& \multicolumn{3}{c|}{\textbf{NeighborMean}} 
& \multicolumn{3}{c|}{\textbf{Feat. Prop.}} 
& \multicolumn{3}{c|}{\textbf{TAPE}} 
& \multicolumn{3}{c|}{\textbf{TA\_E}} 
& \multicolumn{3}{c|}{\textbf{LLM4NG}} 
& \multicolumn{3}{c}{\textbf{Ours}} \\
\cline{2-22}
& Cora & Pub. & Arxiv 
& Cora & Pub. & Arxiv 
& Cora & Pub. & Arxiv 
& Cora & Pub. & Arxiv 
& Cora & Pub. & Arxiv 
& Cora & Pub. & Arxiv 
& Cora & Pub. & Arxiv \\
\hline
0.00      & 88.54 & \textit{87.99} & \ul{88.91} & 88.17 & \textit{87.16} & \ul{88.25} & 88.17 & \textit{87.16} & \ul{88.25} & 84.49 & \textit{83.86} & \ul{83.90} & 83.36 & \textit{82.74} & \ul{82.59} & 88.06 & \textit{88.47} & \ul{87.06} & \textbf{89.39} & \textit{\textbf{90.17}} & \ul{\textbf{90.39}} \\
0.33      & 87.39 & \textit{87.06} & \ul{88.91} & 86.36 & \textit{85.15} & \ul{86.84} & 86.36 & \textit{84.99} & \ul{86.84} & 84.53 & \textit{84.27} & \ul{83.96} & 83.32 & \textit{83.03} & \ul{80.93} & 86.21 & \textit{87.06} & \ul{86.62} & \textbf{88.98} & \textit{\textbf{89.17}} & \ul{\textbf{89.46}} \\
0.50      & 84.81 & \textit{82.64} & \ul{84.77} & 83.37 & \textit{82.52} & \ul{83.65} & 83.58 & \textit{82.47} & \ul{83.73} & 81.74 & \textit{80.27} & \ul{81.46} & 80.59 & \textit{79.39} & \ul{80.65} & 84.07 & \textit{84.19} & \ul{83.82} & \textbf{86.01} & \textit{\textbf{86.66}} & \ul{\textbf{86.69}} \\
0.66      & 84.55 & \textit{83.66} & \ul{84.36} & 85.63 & \textit{83.70} & \ul{85.03} & 85.63 & \textit{83.70} & \ul{85.03} & 85.36 & \textit{84.01} & \ul{85.01} & 83.62 & \textit{81.52} & \ul{83.33} & 84.33 & \textit{83.18} & \ul{82.74} & \textbf{87.13} & \textit{\textbf{86.88}} & \ul{\textbf{86.73}} \\
0.83      & 83.50 & \textit{81.22} & \ul{83.87} & 82.56 & \textit{81.91} & \ul{82.69} & 82.80 & \textit{81.95} & \ul{82.94} & 80.49 & \textit{79.52} & \ul{79.17} & 78.58 & \textit{77.90} & \ul{77.30} & 83.49 & \textit{83.29} & \ul{82.85} & \textbf{85.33} & \textit{\textbf{85.35}} & \ul{\textbf{85.78}} \\
0.90      & 76.51 & \textit{72.13} & \ul{76.74} & 74.03 & \textit{73.25} & \ul{73.70} & 74.74 & \textit{73.61} & \ul{74.72} & 74.89 & \textit{72.59} & \ul{74.62} & 73.55 & \textit{71.46} & \ul{72.76} & 76.38 & \textit{76.95} & \ul{76.79} & \textbf{78.80} & \textit{\textbf{79.37}} & \ul{\textbf{79.40}} \\
1.00      & 78.38 & \textit{75.30} & \ul{78.34} & 75.54 & \textit{74.89} & \ul{73.94} & 76.16 & \textit{75.32} & \ul{74.42} & 76.99 & \textit{77.18} & \ul{77.27} & 75.89 & \textit{74.69} & \ul{75.19} & 80.58 & \textit{80.95} & \ul{81.03} & \textbf{83.00} & \textit{\textbf{82.65}} & \ul{\textbf{83.32}} \\
1.16      & 79.19 & \textit{76.79} & \ul{78.47} & 79.97 & \textit{78.75} & \ul{78.98} & 80.00 & \textit{78.63} & \ul{78.97} & 80.31 & \textit{78.69} & \ul{80.16} & 78.00 & \textit{76.24} & \ul{78.60} & 80.77 & \textit{80.11} & \ul{79.33} & \textbf{83.21} & \textit{\textbf{83.06}} & \ul{\textbf{83.03}} \\
1.23      & 73.02 & \textit{66.44} & \ul{72.45} & 68.79 & \textit{68.83} & \ul{69.62} & 69.79 & \textit{69.31} & \ul{71.56} & 71.16 & \textit{70.32} & \ul{73.53} & 69.22 & \textit{68.73} & \ul{71.05} & 76.13 & \textit{75.83} & \ul{76.25} & \textbf{78.37} & \textit{\textbf{78.35}} & \ul{\textbf{78.16}} \\
1.33      & 77.61 & \textit{73.78} & \ul{77.00} & 76.27 & \textit{74.67} & \ul{74.07} & 76.05 & \textit{75.06} & \ul{74.35} & 76.42 & \textit{76.19} & \ul{74.00} & 74.72 & \textit{75.19} & \ul{70.16} & 80.37 & \textit{81.08} & \ul{79.25} & \textbf{82.58} & \textit{\textbf{82.69}} & \ul{\textbf{82.42}} \\
1.40      & 66.26 & \textit{61.46} & \ul{66.21} & 60.71 & \textit{60.43} & \ul{59.27} & 62.48 & \textit{62.62} & \ul{61.79} & 69.54 & \textit{68.71} & \ul{68.87} & 67.18 & \textit{64.57} & \ul{65.61} & 73.02 & \textit{72.98} & \ul{73.03} & \textbf{75.24} & \textit{\textbf{75.87}} & \ul{\textbf{75.64}} \\
1.50      & 69.63 & \textit{64.37} & \ul{68.22} & 62.98 & \textit{62.82} & \ul{59.84} & 64.11 & \textit{63.31} & \ul{61.85} & 71.59 & \textit{74.52} & \ul{74.89} & 67.04 & \textit{69.78} & \ul{72.37} & 77.19 & \textit{76.07} & \ul{76.59} & \textbf{80.30} & \textit{\textbf{80.30}} & \ul{\textbf{78.96}} \\
1.56      & 64.36 & \textit{59.31} & \ul{64.68} & 63.19 & \textit{64.41} & \ul{64.94} & 64.59 & \textit{65.43} & \ul{67.20} & 70.93 & \textit{69.77} & \ul{71.84} & 67.08 & \textit{64.36} & \ul{67.60} & 73.57 & \textit{73.55} & \ul{73.29} & \textbf{76.23} & \textit{\textbf{76.39}} & \ul{\textbf{76.78}} \\
1.66      & 71.71 & \textit{66.45} & \ul{70.14} & 72.01 & \textit{70.45} & \ul{69.60} & 71.94 & \textit{69.94} & \ul{71.19} & 73.50 & \textit{72.62} & \ul{74.02} & 70.82 & \textit{71.60} & \ul{73.61} & 77.50 & \textit{76.79} & \ul{75.44} & \textbf{78.97} & \textit{\textbf{80.06}} & \ul{\textbf{78.69}} \\
1.73      & 62.99 & \textit{57.49} & \ul{62.55} & 56.51 & \textit{56.59} & \ul{55.63} & 58.11 & \textit{58.17} & \ul{57.75} & 66.86 & \textit{68.67} & \ul{68.91} & 65.77 & \textit{66.51} & \ul{66.40} & 72.18 & \textit{71.79} & \ul{71.61} & \textbf{74.13} & \textit{\textbf{75.55}} & \ul{\textbf{75.12}} \\
1.80      & 47.56 & \textit{44.66} & \ul{47.45} & 42.36 & \textit{45.41} & \ul{44.58} & 43.36 & \textit{46.11} & \ul{45.85} & 59.05 & \textit{57.92} & \ul{57.27} & 59.92 & \textit{49.49} & \ul{56.93} & 62.19 & \textit{63.26} & \ul{63.83} & \textbf{65.38} & \textit{\textbf{65.44}} & \ul{\textbf{64.34}} \\
1.83      & 69.70 & \textit{66.22} & \ul{69.78} & 64.52 & \textit{63.63} & \ul{59.68} & 63.87 & \textit{63.06} & \ul{63.63} & 67.11 & \textit{72.63} & \ul{69.37} & 61.41 & \textit{72.22} & \ul{64.81} & 77.26 & \textit{78.67} & \ul{75.70} & \textbf{79.70} & \textit{\textbf{80.74}} & \ul{\textbf{79.85}} \\
1.90      & 55.06 & \textit{51.73} & \ul{54.37} & 44.36 & \textit{46.72} & \ul{43.74} & 49.14 & \textit{46.94} & \ul{46.30} & 62.51 & \textit{65.28} & \ul{67.35} & 58.15 & \textit{59.51} & \ul{63.36} & 69.78 & \textit{69.85} & \ul{69.28} & \textbf{72.37} & \textit{\textbf{71.53}} & \ul{\textbf{72.79}} \\
2.06      & 54.93 & \textit{48.75} & \ul{54.47} & 50.17 & \textit{50.81} & \ul{50.01} & 52.22 & \textit{52.37} & \ul{52.55} & 64.16 & \textit{64.27} & \ul{63.88} & 61.32 & \textit{60.52} & \ul{59.86} & 70.00 & \textit{69.23} & \ul{68.14} & \textbf{72.83} & \textit{\textbf{72.38}} & \ul{\textbf{71.97}} \\
2.13      & 41.63 & \textit{42.26} & \ul{42.39} & 39.02 & \textit{40.60} & \ul{41.01} & 39.74 & \textit{40.87} & \ul{41.05} & 58.44 & \textit{60.78} & \ul{60.37} & 60.12 & \textit{63.94} & \ul{59.28} & 60.51 & \textit{61.96} & \ul{60.75} & \textbf{64.11} & \textit{\textbf{64.40}} & \ul{\textbf{65.93}} \\
2.16      & 62.29 & \textit{57.56} & \ul{60.15} & 58.06 & \textit{57.42} & \ul{55.40} & 55.89 & \textit{55.08} & \ul{55.40} & 65.37 & \textit{67.82} & \ul{64.59} & 64.00 & \textit{68.96} & \ul{60.07} & 72.07 & \textit{72.81} & \ul{69.33} & \textbf{76.74} & \textit{\textbf{75.26}} & \ul{\textbf{74.00}} \\
2.23      & 53.31 & \textit{48.45} & \ul{51.43} & 41.93 & \textit{40.82} & \ul{41.02} & 45.39 & \textit{43.63} & \ul{42.85} & 62.90 & \textit{64.57} & \ul{65.83} & 61.78 & \textit{61.06} & \ul{63.93} & 69.26 & \textit{68.62} & \ul{66.91} & \textbf{72.67} & \textit{\textbf{72.57}} & \ul{\textbf{70.84}} \\
2.30      & 35.85 & \textit{37.88} & \ul{36.74} & 35.49 & \textit{35.27} & \ul{35.48} & 34.37 & \textit{36.57} & \ul{35.75} & 57.22 & \textit{51.53} & \ul{59.80} & 54.81 & \textit{48.82} & \ul{57.95} & 58.34 & \textit{59.73} & \ul{59.46} & \textbf{62.02} & \textit{\textbf{62.82}} & \ul{\textbf{62.94}} \\
2.46      & 35.23 & \textit{35.95} & \ul{34.64} & 35.62 & \textit{35.47} & \ul{34.18} & 36.00 & \textit{35.86} & \ul{33.54} & 51.54 & \textit{53.49} & \ul{55.88} & 51.01 & \textit{49.45} & \ul{52.37} & 59.46 & \textit{59.22} & \ul{56.40} & \textbf{63.69} & \textit{\textbf{65.79}} & \ul{\textbf{63.20}} \\
2.56      & 45.85 & \textit{40.81} & \ul{45.53} & 35.67 & \textit{35.89} & \ul{34.28} & 38.93 & \textit{38.67} & \ul{38.34} & 59.01 & \textit{61.76} & \ul{56.05} & 56.96 & \textit{60.74} & \ul{53.58} & 64.27 & \textit{65.56} & \ul{62.30} & \textbf{68.82} & \textit{\textbf{68.79}} & \ul{\textbf{67.68}} \\
2.63      & 33.66 & \textit{34.47} & \ul{33.78} & 29.58 & \textit{29.66} & \ul{29.83} & 30.09 & \textit{30.18} & \ul{29.34} & 55.63 & \textit{55.91} & \ul{55.94} & 55.11 & \textit{51.38} & \ul{53.46} & 59.33 & \textit{58.64} & \ul{57.46} & \textbf{63.19} & \textit{\textbf{63.06}} & \ul{\textbf{64.07}} \\
2.70      & 23.70 & \textit{29.26} & \ul{23.70} & 35.51 & \textit{35.92} & \ul{35.51} & 35.51 & \textit{36.33} & \ul{35.51} & 50.18 & \textit{42.04} & \ul{37.03} & 51.85 & \textit{34.45} & \ul{33.33} & 49.26 & \textit{48.89} & \ul{51.85} & \textbf{53.33} & \textit{\textbf{56.67}} & \ul{\textbf{56.30}} \\
2.96      & 30.52 & \textit{32.45} & \ul{29.88} & 28.05 & \textit{29.00} & \ul{26.65} & 28.48 & \textit{29.70} & \ul{25.89} & 49.96 & \textit{50.81} & \ul{47.87} & 47.41 & \textit{48.00} & \ul{47.58} & 54.30 & \textit{55.28} & \ul{53.26} & \textbf{60.57} & \textit{\textbf{61.73}} & \ul{\textbf{59.70}} \\
3.03      & 22.22 & \textit{30.37} & \ul{25.56} & 29.39 & \textit{29.39} & \ul{28.98} & 29.79 & \textit{29.39} & \ul{28.57} & 51.66 & \textit{40.74} & \ul{41.66} & 53.33 & \textit{42.59} & \ul{38.89} & 50.00 & \textit{45.19} & \ul{48.15} & \textbf{57.04} & \textit{\textbf{55.56}} & \ul{\textbf{61.11}} \\
3.36      & 19.63 & \textit{29.63} & \ul{19.63} & 29.39 & \textit{29.39} & \ul{28.57} & 29.39 & \textit{29.39} & \ul{28.57} & 40.37 & \textit{40.00} & \ul{41.66} & 40.74 & \textit{39.26} & \ul{38.89} & 44.44 & \textit{47.41} & \ul{45.19} & \textbf{54.81} & \textit{\textbf{54.07}} & \ul{\textbf{54.81}} \\ \hline
\end{tabular}}
\end{minipage}

\end{table*}

\subsection{Retrieval-Refined Contrastive Learning}
R2CL module is designed to periodically inject retrieval-guided semantic refinements into the graph structure and node embeddings. Modifications on both node embeddings and graph topology encourage learned representations to become semantically distinctive while being label-consistent, thereby improving discriminability.

Given the enriched graph $\widetilde{\mathcal{G}} = (\widetilde{\mathcal{V}}, \widetilde{\mathcal{E}}, \widetilde{\mathcal{X}})$ produced by SGGM, 
R2CL regularizes node representations through two contrastive views.
At every $T$ epochs, retrieval-refined augmentation on $\widetilde{\mathcal{G}}$ is performed. 
Specifically, R2CL maintains the embedding store 
$\mathcal{M}$ dynamic, where each $\mathbf{e}_i\in\mathbb{R}^d$ is the current representation of node $v_i$. 
For each randomly selected anchor node $v_i$, its top-$K$ nearest neighbors \(\mathcal{N}_i\) are retrieved from $\mathcal{M}$ based on similarity \(\frac{\mathbf{e}_i^\top \mathbf{e}_j}{\|\mathbf{e}_i\| \|\mathbf{e}_j\|} \). $\mathcal{N}_i$ is split into same-class and cross-class sets $\mathcal{S}_i$ and $\mathcal{D}_i$ according to labels $\{y_j\}$. 
The anchor text $t_i$ is then semantically optimized by $f_{\text{LLM}}$ for embedding-level modification, conditioned on the retrieved contexts $\{t_j : j \in \mathcal{S}_i \cup \mathcal{D}_i\}$, illustrated by Figure~\ref{Contrastive_learning_prompt} (i):
\begin{equation}
\hat{t}_i = f_{\text{LLM}}\big(t_i;\{t_j:j\in\mathcal{S}_i\cup\mathcal{D}_i\}\big),
\qquad
\hat{\mathbf{e}}_i = f_{\text{emb}}(\hat{t}_i).
\end{equation}

Simultaneously, the LLM determines whether an edge should exist between the anchor $v_i$ and each retrieved node $v_j$. Applying this refinement to all anchors yields a retrieval-refined graph 
$\widetilde{\mathcal{G}}^{\text{RAG}}$ 
with updated feature matrix $\mathbf{X}^{\text{RAG}}$ and adjacency matrix $\mathbf{A}^{\text{RAG}}$.

To perform contrastive learning, R2CL constructs two graph views. 
The first view $\widetilde{\mathcal{G}}^{(1)}$ is generated by applying random edge dropping and random feature masking to $\widetilde{\mathcal{G}}$, 
producing $(\mathbf{X}^{(1)},\mathbf{A}^{(1)})$. 
The second view $\widetilde{\mathcal{G}}^{(2)}$ is generated from $\widetilde{\mathcal{G}}^{\text{RAG}}$ when the current epoch triggers refinement, 
or otherwise from another stochastically perturbed copy of $\widetilde{\mathcal{G}}$, 
producing $(\mathbf{X}^{(2)},\mathbf{A}^{(2)})$. Let $f_\theta$ denote a GCN encoder and $g_\phi$ a projection head. For each node $v_i$ in a batch $\mathcal{B}$, we compute normalized embeddings from both views:
\begin{equation}
\mathbf{h}_i^{(k)} = f_\theta(\mathbf{X}^{(k)},\mathbf{A}^{(k)})_i,\quad
\mathbf{z}_i^{(k)} = \frac{g_\phi(\mathbf{h}_i^{(k)})}{\|g_\phi(\mathbf{h}_i^{(k)})\|_2},
\quad k\in\{1,2\}.
\end{equation}

We concatenate the embeddings from the two views and optimize them with the supervised contrastive objective \(\mathcal{L}\)
:
% \begin{equation}
% \mathcal{L}
% =
% \frac{1}{|\mathcal{B}|}
% \sum_{i\in\mathcal{B}}
% \frac{-1}{|P(i)|}
% \sum_{p\in P(i)}
% \log
% \frac{\exp(\mathbf{z}_i\cdot\mathbf{z}_p/\tau_{temp})}
% {\sum_{a\in A(i)}\exp(\mathbf{z}_i\cdot\mathbf{z}_a/\tau_{temp})},
% \end{equation}
\begin{equation}
\mathcal{L}
=
\frac{1}{|\mathcal{B}|}
\sum_{i\in\mathcal{B}}
\omega \cdot
\frac{-1}{|P(i)|}
\sum_{p\in P(i)}
\log
\frac{\exp(\mathbf{z}_i \cdot \mathbf{z}_p / \tau_{temp})}
{\sum_{a\in A(i)} \exp(\mathbf{z}_i \cdot \mathbf{z}_a / \tau_{temp})},
\end{equation}

where $P(i)=\{p\neq i\mid y_p=y_i\}$, $A(i)=\{a\neq i\}$, and $\tau_{temp}$ is the temperature hyperparameter. 
To strengthen the contribution of generated nodes $\mathcal{V}=V\cup\hat{V}$, their loss multiplicative weight $\omega>1$.

The final robust representations and the LLM-refined enriched graph are subsequently fed into the downstream GNN classifier for node classification, trained in the presence of injected attacks.

Unlike previous contrastive learning approaches in SimGCL~\cite{liu2024simgcl} and SGL~\cite{wu2021self}, which rely on structural perturbations for contrastive views, R2CL conducts contrastive optimization on LLM-refined retrieval-guided views during augmentation before GNN classification. Section~\ref{contrastive_comparison_section} further demonstrates its superiority.

\begin{figure*}[t]
  \centering
  \includegraphics[width=\linewidth,height=3cm]{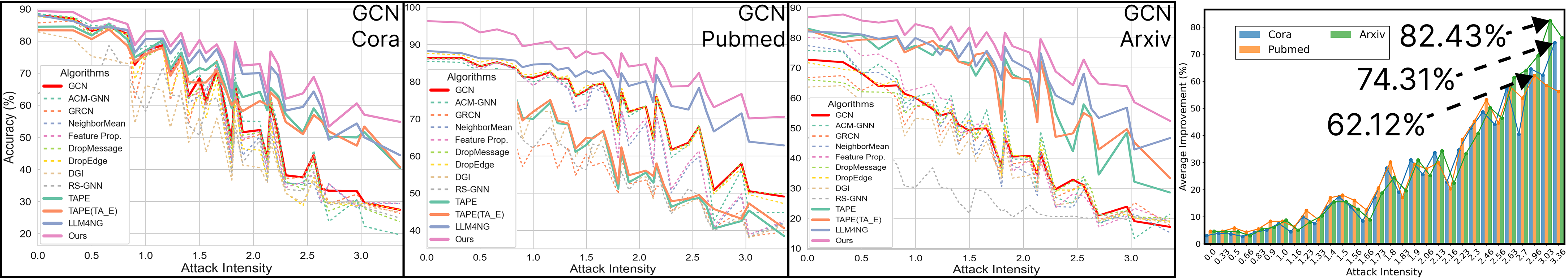}
  \caption{Accuracy vs. attack intensity (left). Relative improvement over selected well-performing baselines in Table~\ref{Main_Table}.b (right).}
  \label{visualization}
\end{figure*}

\section{Experiments}
In this section, we study the following research questions:

\textbf{RQ1}: Does RoGRAD outperform all baselines on both non-deficient and deficient graphs, particularly under low-to-moderate attacks?

\textbf{RQ2}: How does each component contribute to the overall performance of RoGRAD?

\textbf{RQ3}: Can RoGRAD build the strongest robustness under compound graph deficiencies, allowing LLM-as-Enhancer to surpass previously more robust non-LLM baselines?

\textbf{RQ4}: Can R2CL, by introducing LLM guided modification, outperform prior graph contrastive learning approaches?

\subsection{Experimental Setup}
We conducted experiments on the \textsc{Cora}~\cite{yang2016revisiting}, \textsc{PubMed}~\cite{yang2016revisiting}, and \textsc{Arxiv}~\cite{hu2020open} datasets (see Appendix Section~\ref{datasets} for details). 
In addition to the configurations in Section~\ref{experiment_settings}, we further specify the following settings. For the SGGM, we retrieve $k=10$ same-category exemplars and apply similarity thresholds of 0.85 ($r_i$), 0.6 ($a_i$), 0.3 ($o_i$), and 0.7 ($d_i$). Generated nodes were linked when similarity $>\tau=0.7$. For the R2CL module, a 4-layer GCN (256-dim) with a 128-dim projection head is adopted as encoder, batch size = 128, $\tau_{temp}=0.07$, and loss weight $\omega=2.0$. It is trained for 50 epochs, applying RAG enhancement every $T=5$ epochs using 15 anchors (3 same + 7 different) with edge drop = 0.1 and feature mask = 0.1.

\subsection{Results and Analysis}
\subsubsection{Performance under Non-deficient and Deficient Graphs (RQ1)}
Experiments demonstrate RoGRAD's (Ours) consistent advantages over conventional GNN-based and LLM-enhanced baselines on both clean and deficient graphs. As shown in Table~\ref{Main_Table}.a, under clean graphs, it achieves the highest accuracy on all three datasets and GNN backbones. For instance, on PubMed-SAGE, RoGRAD reaches the highest 97.04\% accuracy, outperforming the best baseline (LLM4NG) by 7.28\% and 9.09\% for the top-4 best-performing baselines. RoGRAD also demonstrates considerable improvement, up to 10.56\% improvement on other datasets and GNN architectures.

According to Table~\ref{Main_Table}.b, RoGRAD consistently achieves the highest accuracy at every attack intensity on deficient graphs. Figure~\ref{visualization} shows that as attack intensity increases, LLM-as-Enhancer baselines outperform GNN-based ones, but still lag behind RoGRAD (pink). RoGRAD achieves improvement up to 62.12\% on PubMed, 74.31\% on Cora, and 82.43\% on Arxiv over best-performing baselines.

Under low-to-moderate attacks, RoGRAD successfully enables LLM-as-Enhancer frameworks to surpass all non-LLM baselines. The observed reversal highlights the effectiveness of RoGRAD’s retrieval-augmented generation and contrastive refinement in mitigating the drawbacks of previously LLM-based augmentation.

\subsubsection{Ablation Study (RQ2)}
\begin{figure}[t]
  \centering
  
  \includegraphics[width=\linewidth,height=5.5cm]{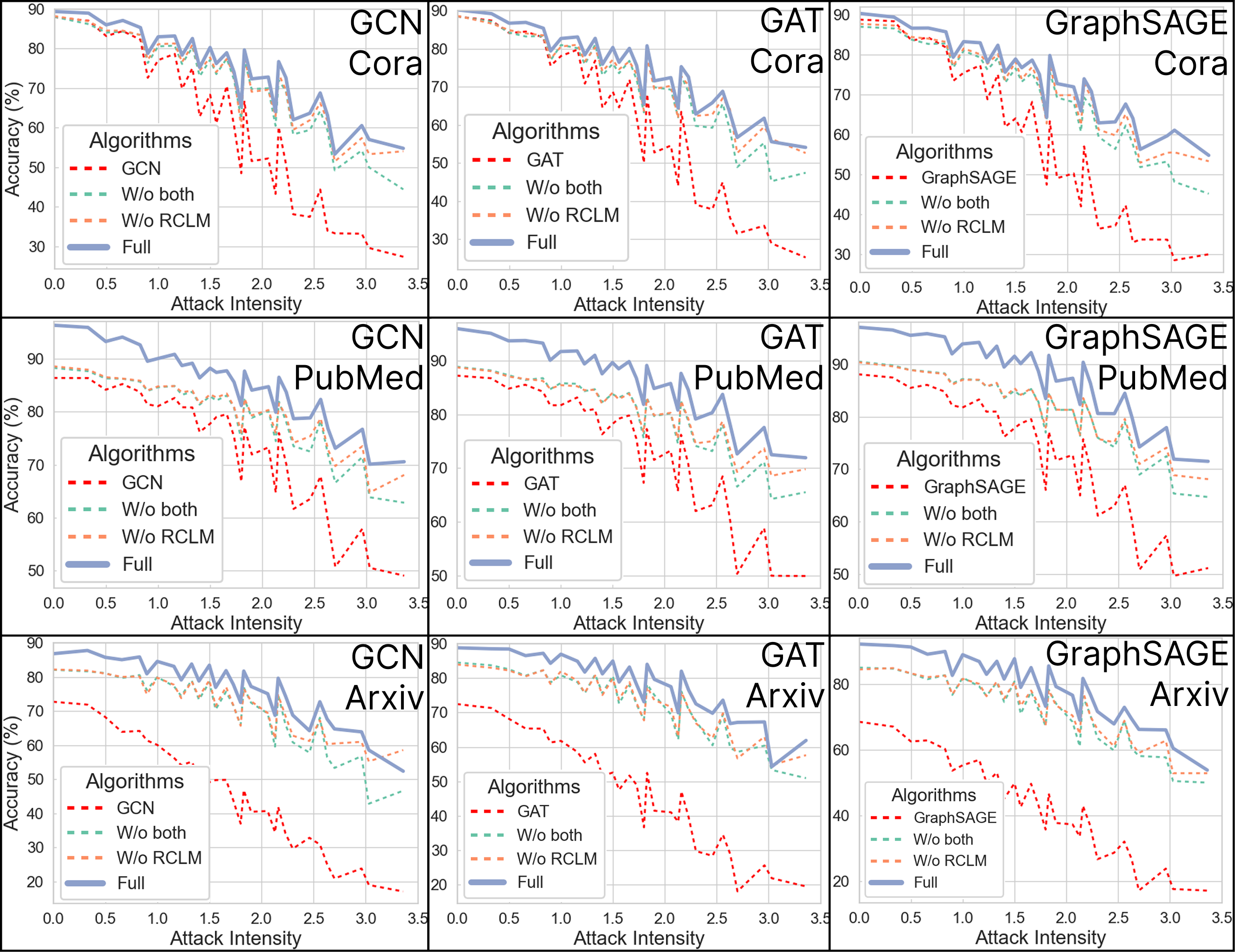}
  \caption{Ablations on Cora, PubMed, and Arxiv datasets.}
  \label{Ablation}
\end{figure}

Figure~\ref{Ablation} confirms the contribution of each module. Comparing W/o Both with base GNNs underscores that graph enrichment in Section~\ref{enrichment} leads to much higher accuracy by supplementing missing information. The gap between W/o Both and W/o RCLM demonstrates the effect of SGGM, which becomes increasingly evident under higher attack intensities, reaching more than 10\% on Cora. Although this gap varies across datasets and GNN backbones, it is also clear on PubMed and Arxiv. SGGM is proven to yield further gains at high attack levels by injecting diverse, class-consistent information. Finally, the improvement of Full over W/o RCLM highlights the indispensable role of RCLM, which significantly enhances accuracy under severe deficiencies through retrieval-refined contrastive optimization. These results show that each module contributes unique value, and their integration allows RoGRAD to consistently achieve the best accuracy across datasets.

\begin{figure}[b]
  \centering
  \includegraphics[width=\linewidth,height=3cm]{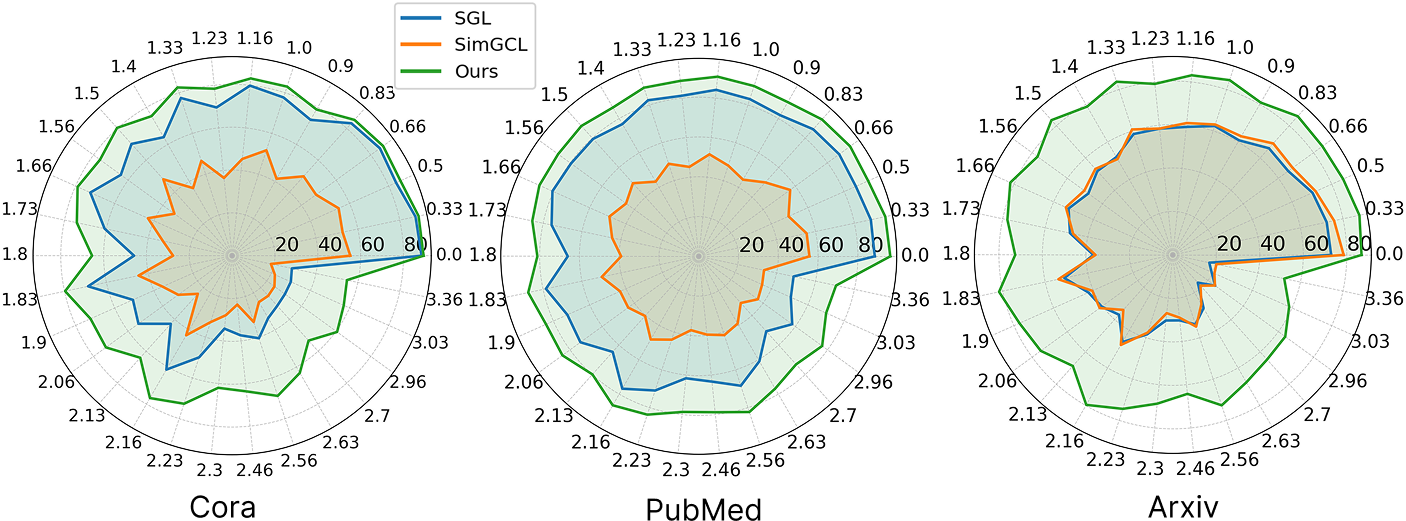}
  \caption{Accuracy across datasets: Ours vs. GCL baselines.}
  \label{contrastive_comparison}
\end{figure}

\subsubsection{Robustness under Compound Deficiencies (RQ3)}

In Table~\ref{robustness}, RoGRAD establishes the strongest robustness across all datasets. On PubMed, it reaches 96.29\% clean accuracy and an outstanding 90.64\% worst-case accuracy, far higher than the strongest baseline LLM4NG at 84.48\%. On Arxiv, RoGRAD improves average accuracy to 84.99\%, surpassing TAPE (80.01\%) and LLM4NG (80.12\%). On Cora, RoGRAD attains the highest average accuracy, 85.68\%, and Norm-AUC 0.86. These results show that, unlike prior LLM-based methods that trail behind robust GNNs, RoGRAD surpasses both, proving most robust under compound deficiencies.

\begin{table}[h]
\centering
\caption{Robustness across all benchmark datasets on GCN.}
\label{robustness}
\large
\renewcommand{\arraystretch}{1.2}
\resizebox{\linewidth}{!}{%
\begin{tabular}{l|ccc|ccc|ccc|ccc}
\toprule
\multirow{2}{*}{Algorithm} 
 & \multicolumn{3}{c|}{Clean Acc} 
 & \multicolumn{3}{c|}{Worst Acc} 
 & \multicolumn{3}{c|}{Avg Acc} 
 & \multicolumn{3}{c}{Norm-AUC} \\
 
 & Cora & Pub. & Arx. 
 & Cora & Pub. & Arx.
 & Cora & Pub. & Arx. 
 & Cora & Pub. & Arx. \\
\midrule
GCN          & 87.98 & 86.36 & 72.73 & 75.59 & 82.33 & 62.03 & 82.58 & 84.46 & 67.98 & 0.83 & 0.85 & 0.69 \\
DropEdge     & 87.61 & 87.60 & 71.73 & 71.99 & 82.91 & 61.17 & 81.35 & 85.41 & 67.29 & 0.83 & 0.86 & 0.68 \\
DropMsg      & 87.65 & 86.21 & 66.04 & 76.37 & 82.26 & 56.63 & 82.52 & 84.38 & 61.82 & 0.83 & 0.85 & 0.62 \\
Feat. Prop.  & 88.17 & 86.09 & 80.14 & 77.46 & 82.39 & 65.33 & 83.30 & 84.20 & 73.58 & 0.84 & 0.84 & 0.74 \\
Nbr. Mean    & 88.17 & 86.10 & 77.39 & 76.93 & 82.19 & 58.86 & 83.07 & 84.10 & 68.85 & 0.84 & 0.84 & 0.70 \\
ACM-GNN      & 88.54 & 85.45 & 75.77 & 78.52 & 82.11 & 61.48 & 84.17 & 83.63 & 68.93 & 0.85 & 0.84 & 0.70 \\
DGI          & 82.88 & 80.97 & 63.61 & 59.38 & 67.95 & 50.44 & 72.93 & 75.68 & 58.08 & 0.74 & 0.77 & 0.59 \\
RS-GNN       & 63.59 & 72.97 & 46.73 & 61.12 & 66.52 & 34.34 & 65.54 & 69.58 & 40.54 & 0.65 & 0.70 & 0.41 \\
GRCN         & 85.66 & 65.72 & 66.74 & 67.42 & 58.62 & 54.63 & 78.64 & 63.20 & 61.68 & 0.80 & 0.64 & 0.63 \\
TAPE         & 84.49 & 82.14 & 83.04 & 77.16 & 73.03 & 76.13 & 81.48 & 77.87 & 80.01 & 0.82 & 0.78 & 0.80 \\
TA\_E        & 83.36 & 81.43 & 82.61 & 75.47 & 72.43 & 76.01 & 80.23 & 77.60 & 79.34 & 0.81 & 0.78 & 0.80 \\
LLM4NG       & 88.06 & 88.25 & 82.09 & 78.29 & 84.48 & 77.02 & 83.68 & 86.44 & 80.12 & 0.84 & 0.87 & 0.81 \\
Ours         & \cellcolor{gray!25}\textbf{89.39} & \cellcolor{gray!25}\textbf{96.29} & \cellcolor{gray!25}\textbf{86.79} & \cellcolor{gray!25}\textbf{80.88} & \cellcolor{gray!25}\textbf{90.64} & \cellcolor{gray!25}\textbf{81.94} & \cellcolor{gray!25}\textbf{85.68} & \cellcolor{gray!25}\textbf{93.61} & \cellcolor{gray!25}\textbf{84.99} & \cellcolor{gray!25}\textbf{0.86} & \cellcolor{gray!25}\textbf{0.94} & \cellcolor{gray!25}\textbf{0.85} \\
\bottomrule
\end{tabular}}
\end{table}

\subsubsection{Robust Contrastive Learning with R2CL (RQ4)}\label{contrastive_comparison_section}
Traditional contrastive learning methods for graphs, such as SimGCL~\cite{liu2024simgcl} and SGL~\cite{wu2021self}, rely on one-shot perturbations to construct views, which offer limited guidance to iteratively preserve discriminative consistency. In contrast, R2CL continuously optimizes graph views through LLM-perturbed semantic refinements and structures to utilize LLM’s unique strength in generating. As illustrated in Figure~\ref{contrastive_comparison}, R2CL consistently encloses a larger shaded performance region than SimGCL and SGL, reflecting higher accuracy and stronger robustness across a wide range of attack intensities. The iterative refinement explicitly enforces intra-class alignment and inter-class separation. These results confirm that R2CL not only surpasses existing graph contrastive baselines but also establishes robust advantages of iterative, LLM-refined contrastive learning.

\subsubsection{Effect of Hyperparameter $\lambda$}
\begin{figure}[h]
  \centering
  \includegraphics[width=\linewidth]{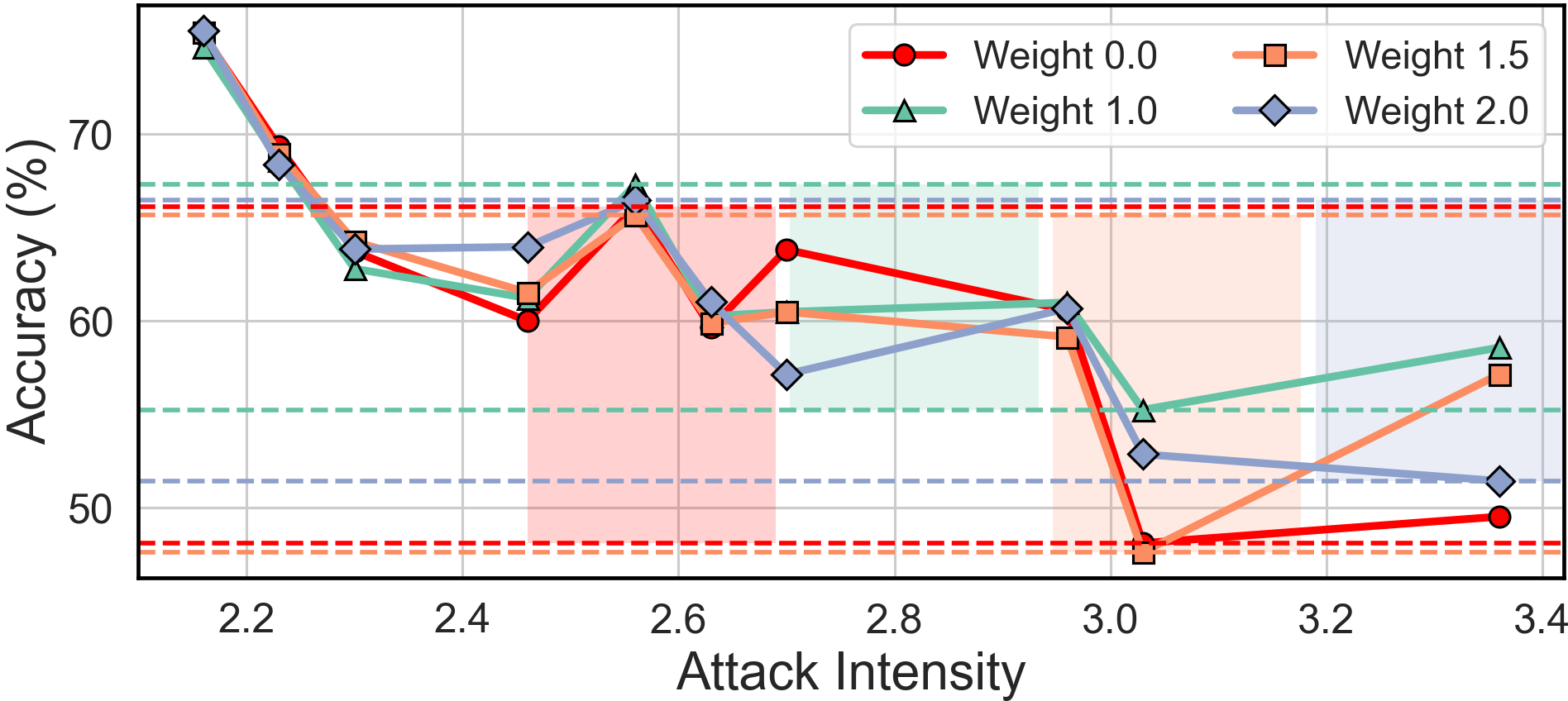}
  \caption{Effect of keyword weight $\lambda$ on GCN accuracy (Arxiv).}
  \label{keywords_weight}
\end{figure}

The hyperparameter $\lambda$, which denotes the fusion weight of \textit{Main Textual Content: Keywords}, controls the weight of keywords in the semantic-guided generation, directly influencing the balance between main content and class-specific terminology. As illustrated in Figure~\ref{keywords_weight}, different settings of $\lambda$ yield distinct accuracy trends under varying attack intensities.

When $\lambda$ = 0.0, GCN remains relatively stable under mild attacks but deteriorates rapidly as attack intensity increases. Moderate setting when $\lambda$ = 1 yields the most robust performance by reinforcing intra-class alignment, but its performance under mild attacks is not sufficient, whereas $\lambda$ = 1.5 presents the largest fluctuation. Overall, moderate discrete choices of $\lambda$ = 2 provide balanced performance and robustness to integrate keywords for class alignment.

\section{Conclusion}
In this work, we conducted the first comprehensive evaluation of LLM-on-graph learning under compound graph deficiencies. Our study reveals that existing LLM-as-Enhancer paradigms often deliver unstable performance and may fail to surpass traditional GNN-based methods. This finding challenges the common assumption that LLMs are generally better GNN helpers. Beyond empirical gains, our work introduces a new iterative refinement paradigm RoGRAD, the first retrieval-augmented generation framework that targets robust graph learning under graph deficiencies to refine and stabilize LLM guidance. RoGRAD leverages retrieval-augmented refinement of LLM guidance to enhance intra-class consistency and inter-class separation. At the same time, it integrates supplementary information to address graph deficiencies, thereby establishing a principled mechanism for generating reliable and robust node representations. Extensive experiments demonstrate that RoGRAD sets a new benchmark for robust graph learning with LLMs. These results establish RoGRAD as a new robust Web graph learning framework and provide the first evidence that retrieval-augmented iterative refinement is crucial for integrating LLMs on graphs.

\clearpage
\bibliographystyle{ACM-Reference-Format}
\bibliography{sample-base}

\appendix

% \section*{Relevance Declaration \& Code Availablity}
% This work is relevant to the Web as it studies algorithms and analysis for \textbf{incomplete, noisy, and partially observed Web-related graphs}, such as citation and social networks. The paper perfectly aligns with the track topics on graph neural networks and the integration of \textbf{large language models} for \textbf{Web-related graphs}.

% \textbf{Our code is available at:} 

% https://anonymous.4open.science/r/WWW2026-DF70 .

\section{Issues with One-shot LLM Augmentations}\label{Issues}
To illustrate the drawbacks of current LLM-as-Enhancer frameworks, we followed the prompting style used in LLM4NG and TAPE, which ask an LLM to generate academic-style samples in a Title–Abstract format. Below we show two such examples. Despite targeting different domains, the outputs reveal the same issues observed in prior work: (1) semantic homogeneity across categories, where abstracts share nearly identical templates and phrasing, and (2) the one-shot limitation, since low-quality generations cannot be refined. These issues limit the usefulness of LLM-generated augmentations for robust graph learning.

\textbf{Title.} 
\textbf{\underline{Enhancing}} Neural Network Interpretability \textbf{\underline{through}} Layer-wise Relevance Propagation (Class: Neural Networks)

\textbf{Abstract.} 
Neural networks have become \underline{increasingly prevalent} in various domains; \underline{however}, the need for interpretability has emerged as a critical \underline{concern}. \textbf{\underline{This paper introduces}} an innovative framework for enhancing the interpretability of neural networks by leveraging Layer-wise Relevance Propagation (LRP). \underline{We propose an extended version} of LRP that adapts its relevance propagation mechanism to accommodate both feedforward and recurrent architectures, facilitating the analysis of complex neural models. \textbf{\underline{Through extensive experiments}} on various benchmark datasets, we \textbf{\underline{demonstrate}} the effectiveness of our approach in providing meaningful insights into model predictions while maintaining high accuracy. Our framework \textbf{\underline{not only elucidates}} the \underline{decision-making} processes of neural networks \textbf{\underline{but also identifies}} key features influencing those decisions, thus empowering practitioners to understand and trust their models. We further discuss the implications of our findings for areas such as model auditing, and feature engineering. Finally, we outline future research directions aimed at refining interpretability in deep learning architectures.

\textbf{Title.} 
\textbf{\underline{Enhancing}} Sample Efficiency in Reinforcement Learning \textbf{\underline{through}} Modular Exploration (Class: Reinforcement Learning)

\textbf{Abstract.}  
Reinforcement Learning (RL) has achieved \underline{remarkable success} in various domains; \textbf{\underline{however,}} its reliance on extensive data and sample efficiency remains \underline{challenging}. \textbf{\underline{This paper proposes}} a novel approach to improve sample efficiency in RL through the integration of modular exploration strategies. By decomposing the RL learning process into distinct modules, we effectively enable agents to explore environments in a more systematic manner, facilitating faster convergence towards optimal policies. \underline{We introduce a framework} that combines curiosity-driven exploration with episodic memory, allowing agents to prioritize previously unexplored states while retaining valuable experiences from past behaviors. Our empirical evaluations on standard RL benchmarks \textbf{\underline{demonstrate}} significant improvements in learning speed and overall performance relative to traditional exploration methods. Additionally, our approach \textbf{\underline{reveals}} insights into balancing exploration and exploitation through modular components, \textbf{\underline{providing}} a more adaptable and robust solution for continuous and discrete action spaces. The implications of this research extend to multi-agent systems and real-world applications, where sample efficiency is paramount for effective \underline{decision-making}.

% \textbf{Title.} 
% \textbf{\underline{Enhancing}} Rule Learning \textbf{\underline{through}} Adaptive Data Partitioning Techniques (Class: Rule Learning)

% \textbf{Abstract.}
% Rule learning has emerged as a pivotal approach in the field of machine learning, enabling the extraction of interpretable decision rules from complex datasets. \textbf{\underline{However}}, traditional rule learning methods often struggle with high-dimensional data and can result in overfitting or underfitting. \textbf{\underline{This paper introduces}} a novel framework that leverages adaptive data partitioning techniques to optimize the rule learning process. By intelligently partitioning the dataset based on the inherent structure of the data, our method improves the quality of extracted rules while maintaining computational efficiency. We implement several partitioning strategies, including clustering-based and density-based methods, to assess their impact on the performance of rule learning algorithms. Experimental results across various benchmark datasets \textbf{\underline{demonstrate}} a significant increase in accuracy and interpretability of the generated rules when utilizing adaptive partitioning techniques. This research \textbf{\underline{not only enhances}} the effectiveness of rule learning but also \textbf{\underline{provides}} new insights into the data characteristics that influence rule development, paving the way for more robust applications in predictive modeling and decision supports.

\section{Configurations of Baseline Methods}\label{BaseGNNConf}

All base GNNs are implemented with consistent hyperparameter settings for fair comparison. The main configurations are as follows:

\begin{itemize}
    \item \textbf{Shared Parameters.} Hidden dimension = 512, dropout = 0.3, learning rate = 0.01, weight decay (L2 coefficient) = 0.0. Node drop ratio, feature drop ratio, and same-type edge reduction ratio are set to 0.5 unless otherwise specified.
    
    \item \textbf{GCN Encoder.} A 3-layer GCN with hidden size 512 per layer, followed by a linear classifier. Each convolutional layer is followed by ReLU activation and dropout (0.3).
    
    \item \textbf{GAT Encoder.} A 2-layer GAT: the first layer uses 8 attention heads (hidden dimension divided by 8), the second layer uses a single head with hidden dimension 512. A linear classifier projects the final representation.
    
    \item \textbf{GraphSAGE Encoder.} A 2-layer GraphSAGE with hidden size 512, followed by a linear classifier. Each layer is followed by ReLU activation and dropout (0.3).
\end{itemize}

All models output class probabilities through a log-softmax layer and are trained with the Adam optimizer (learning rate = 0.01).
Furthermore, for TAPE and TA\_E baselines, both original node features $h_{\text{orig}}$ and explanatory features $h_{\text{expl}}$ are projected to match the feature dimensionality of the corresponding datasets.

For fair comparison, in Section~\ref{contrastive_comparison_section}, we retain the SGGM and Graph Enrichment components, and replace only the R2CL module with SimGCL and SGL under identical settings.

\section{Research Methods}
Table~\ref{full_table} shows complete results of GCN on the Cora under combinations of deficiencies (NRA, SHA, FDA, SSA). The aggregated performance in the main paper is derived from full tables like this.

\begin{table}[h]
\centering
\caption{Full results GCN Cora}
\label{full_table}
\renewcommand{\arraystretch}{0.6}
\begin{tabular}{ccccccc}
\toprule
NRA & SHA & FDA & SSA & train & test & test\_acc \\
\midrule
0.0 & 0.0 & 0.0 & 0.00 & 0.2 & 0.2 & 87.98 ± 1.45 \\
0.0 & 0.0 & 0.0 & 0.33 & 0.2 & 0.2 & 87.06 ± 1.50 \\
0.0 & 0.0 & 0.0 & 0.66 & 0.2 & 0.2 & 84.44 ± 1.96 \\
0.0 & 0.0 & 0.5 & 0.00 & 0.2 & 0.2 & 86.51 ± 1.32 \\
0.0 & 0.0 & 0.5 & 0.33 & 0.2 & 0.2 & 85.55 ± 1.47 \\
0.0 & 0.0 & 0.5 & 0.66 & 0.2 & 0.2 & 82.51 ± 2.00 \\
0.0 & 0.0 & 0.9 & 0.00 & 0.2 & 0.2 & 84.29 ± 0.88 \\
0.0 & 0.0 & 0.9 & 0.33 & 0.2 & 0.2 & 81.78 ± 0.69 \\
0.0 & 0.0 & 0.9 & 0.66 & 0.2 & 0.2 & 76.23 ± 1.51 \\
0.0 & 0.5 & 0.0 & 0.00 & 0.2 & 0.2 & 82.77 ± 1.90 \\
0.0 & 0.5 & 0.0 & 0.33 & 0.2 & 0.2 & 81.22 ± 1.35 \\
0.0 & 0.5 & 0.0 & 0.66 & 0.2 & 0.2 & 76.93 ± 2.02 \\
0.0 & 0.5 & 0.5 & 0.00 & 0.2 & 0.2 & 79.85 ± 1.37 \\
0.0 & 0.5 & 0.5 & 0.33 & 0.2 & 0.2 & 76.97 ± 2.55 \\
0.0 & 0.5 & 0.5 & 0.66 & 0.2 & 0.2 & 73.05 ± 1.34 \\
0.0 & 0.5 & 0.9 & 0.00 & 0.2 & 0.2 & 72.24 ± 2.04 \\
0.0 & 0.5 & 0.9 & 0.33 & 0.2 & 0.2 & 68.39 ± 1.65 \\
0.0 & 0.5 & 0.9 & 0.66 & 0.2 & 0.2 & 63.51 ± 1.47 \\
0.0 & 0.9 & 0.0 & 0.00 & 0.2 & 0.2 & 70.28 ± 2.05 \\
0.0 & 0.9 & 0.0 & 0.33 & 0.2 & 0.2 & 68.91 ± 1.63 \\
0.0 & 0.9 & 0.0 & 0.66 & 0.2 & 0.2 & 63.07 ± 2.67 \\
0.0 & 0.9 & 0.5 & 0.00 & 0.2 & 0.2 & 59.78 ± 1.11 \\
0.0 & 0.9 & 0.5 & 0.33 & 0.2 & 0.2 & 58.67 ± 1.98 \\
0.0 & 0.9 & 0.5 & 0.66 & 0.2 & 0.2 & 52.42 ± 3.10 \\
0.0 & 0.9 & 0.9 & 0.00 & 0.2 & 0.2 & 43.36 ± 1.87 \\
0.0 & 0.9 & 0.9 & 0.33 & 0.2 & 0.2 & 41.74 ± 1.27 \\
0.0 & 0.9 & 0.9 & 0.66 & 0.2 & 0.2 & 35.08 ± 1.43 \\
0.5 & 0.0 & 0.0 & 0.00 & 0.2 & 0.2 & 80.30 ± 1.08 \\
0.5 & 0.0 & 0.0 & 0.33 & 0.2 & 0.2 & 80.00 ± 2.20 \\
0.5 & 0.0 & 0.0 & 0.66 & 0.2 & 0.2 & 76.59 ± 2.42 \\
0.5 & 0.0 & 0.5 & 0.00 & 0.2 & 0.2 & 76.52 ± 1.51 \\
0.5 & 0.0 & 0.5 & 0.33 & 0.2 & 0.2 & 75.33 ± 2.87 \\
0.5 & 0.0 & 0.5 & 0.66 & 0.2 & 0.2 & 70.44 ± 2.88 \\
0.5 & 0.0 & 0.9 & 0.00 & 0.2 & 0.2 & 68.59 ± 1.37 \\
0.5 & 0.0 & 0.9 & 0.33 & 0.2 & 0.2 & 65.63 ± 3.05 \\
0.5 & 0.0 & 0.9 & 0.66 & 0.2 & 0.2 & 58.81 ± 2.80 \\
0.5 & 0.5 & 0.0 & 0.00 & 0.2 & 0.2 & 74.96 ± 2.67 \\
0.5 & 0.5 & 0.0 & 0.33 & 0.2 & 0.2 & 72.96 ± 1.80 \\
0.5 & 0.5 & 0.0 & 0.66 & 0.2 & 0.2 & 68.37 ± 3.24 \\
0.5 & 0.5 & 0.5 & 0.00 & 0.2 & 0.2 & 68.37 ± 4.15 \\
0.5 & 0.5 & 0.5 & 0.33 & 0.2 & 0.2 & 66.82 ± 2.08 \\
0.5 & 0.5 & 0.5 & 0.66 & 0.2 & 0.2 & 60.22 ± 3.60 \\
0.5 & 0.5 & 0.9 & 0.00 & 0.2 & 0.2 & 57.41 ± 3.39 \\
0.5 & 0.5 & 0.9 & 0.33 & 0.2 & 0.2 & 54.52 ± 2.56 \\
0.5 & 0.5 & 0.9 & 0.66 & 0.2 & 0.2 & 46.82 ± 1.81 \\
0.5 & 0.9 & 0.0 & 0.00 & 0.2 & 0.2 & 64.37 ± 3.63 \\
0.5 & 0.9 & 0.0 & 0.33 & 0.2 & 0.2 & 65.11 ± 1.62 \\
0.5 & 0.9 & 0.0 & 0.66 & 0.2 & 0.2 & 59.11 ± 3.39 \\
0.5 & 0.9 & 0.5 & 0.00 & 0.2 & 0.2 & 50.67 ± 5.61 \\
0.5 & 0.9 & 0.5 & 0.33 & 0.2 & 0.2 & 54.07 ± 1.63 \\
0.5 & 0.9 & 0.5 & 0.66 & 0.2 & 0.2 & 49.71 ± 3.16 \\
0.5 & 0.9 & 0.9 & 0.00 & 0.2 & 0.2 & 36.30 ± 3.33 \\
0.5 & 0.9 & 0.9 & 0.33 & 0.2 & 0.2 & 33.48 ± 2.05 \\
0.5 & 0.9 & 0.9 & 0.66 & 0.2 & 0.2 & 35.26 ± 3.49 \\
0.9 & 0.0 & 0.0 & 0.00 & 0.2 & 0.2 & 63.33 ± 6.02 \\
0.9 & 0.0 & 0.0 & 0.33 & 0.2 & 0.2 & 58.89 ± 6.02 \\
0.9 & 0.0 & 0.0 & 0.66 & 0.2 & 0.2 & 44.07 ± 2.16 \\
0.9 & 0.0 & 0.5 & 0.00 & 0.2 & 0.2 & 49.26 ± 6.48 \\
0.9 & 0.0 & 0.5 & 0.33 & 0.2 & 0.2 & 43.33 ± 10.18 \\
0.9 & 0.0 & 0.5 & 0.66 & 0.2 & 0.2 & 38.52 ± 2.72 \\
0.9 & 0.0 & 0.9 & 0.00 & 0.2 & 0.2 & 41.85 ± 5.32 \\
0.9 & 0.0 & 0.9 & 0.33 & 0.2 & 0.2 & 36.66 ± 7.72 \\
0.9 & 0.0 & 0.9 & 0.66 & 0.2 & 0.2 & 32.22 ± 1.89 \\
0.9 & 0.5 & 0.0 & 0.00 & 0.2 & 0.2 & 62.96 ± 5.62 \\
0.9 & 0.5 & 0.0 & 0.33 & 0.2 & 0.2 & 62.59 ± 4.60 \\
0.9 & 0.5 & 0.0 & 0.66 & 0.2 & 0.2 & 41.11 ± 4.12 \\
0.9 & 0.5 & 0.5 & 0.00 & 0.2 & 0.2 & 46.67 ± 8.72 \\
0.9 & 0.5 & 0.5 & 0.33 & 0.2 & 0.2 & 44.07 ± 8.95 \\
0.9 & 0.5 & 0.5 & 0.66 & 0.2 & 0.2 & 36.67 ± 2.96 \\
0.9 & 0.5 & 0.9 & 0.00 & 0.2 & 0.2 & 34.81 ± 5.90 \\
0.9 & 0.5 & 0.9 & 0.33 & 0.2 & 0.2 & 30.74 ± 6.37 \\
0.9 & 0.5 & 0.9 & 0.66 & 0.2 & 0.2 & 30.37 ± 4.48 \\
0.9 & 0.9 & 0.0 & 0.00 & 0.2 & 0.2 & 60.37 ± 7.55 \\
0.9 & 0.9 & 0.0 & 0.33 & 0.2 & 0.2 & 51.48 ± 5.42 \\
0.9 & 0.9 & 0.0 & 0.66 & 0.2 & 0.2 & 45.19 ± 5.44 \\
0.9 & 0.9 & 0.5 & 0.00 & 0.2 & 0.2 & 43.33 ± 7.82 \\
0.9 & 0.9 & 0.5 & 0.33 & 0.2 & 0.2 & 37.78 ± 4.77 \\
0.9 & 0.9 & 0.5 & 0.66 & 0.2 & 0.2 & 34.07 ± 5.81 \\
0.9 & 0.9 & 0.9 & 0.00 & 0.2 & 0.2 & 33.34 ± 4.38 \\
0.9 & 0.9 & 0.9 & 0.33 & 0.2 & 0.2 & 29.63 ± 5.24 \\
0.9 & 0.9 & 0.9 & 0.66 & 0.2 & 0.2 & 27.41 ± 4.60 \\

\bottomrule
\end{tabular}
\end{table}

\section{Prompts}
The prompt shown below is used for the initial sample generation, where the LLM is instructed to produce a research paper in a given category with its Title and Abstract.

\begin{figure}[h]
  \centering
  \includegraphics[width=\linewidth]{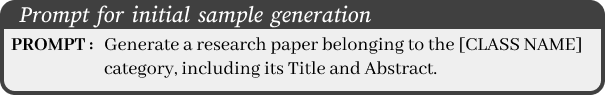}
  \caption{Prompt for initial sample generation.}
  \label{initial_prompt}
\end{figure}

The following subsections present the prompts used in our framework. For brevity, we provide moderately condensed versions that preserve all essential instructions. The complete detailed prompts will be released with the code repository. \textbf{Note:} The following prompts are shown in their \emph{instantiated form} for the PubMed dataset 
(medical research papers in diabetes categories). For other datasets used in our experiments 
(e.g., Cora, Arxiv), we applied the same templates with dataset-specific terminology.

\subsection{Prompt for Initial Generation}
Please generate a medical research paper in the category [\textless category\textgreater], including a title, an abstract, and keywords.  

\textbf{Step 1: Analyze example papers} (\textless similar\_docs\_text\textgreater) to identify:  
-- common terms and methodologies specific to [\textless category\textgreater];  
-- typical clinical research problems in this category;  
-- distinctive approaches separating [\textless category\textgreater] from other diabetes categories.  

\textbf{Step 2: Generate a paper that}  
-- uses \emph{EXACT} medical terms (15--20) from the example papers;  
-- addresses a typical research problem in [\textless category\textgreater];  
-- employs characteristic methodologies of [\textless category\textgreater];  
-- avoids approaches typical of other categories.  

\textbf{Keywords:} 15--20 terms grouped as: Clinical Methodologies (5--7); Therapeutic Approaches (5--7); Biomarkers/Indicators (3--4); Research/Study Types (2--3).  

\textbf{Screening:} Outputs are checked for similarity with [\textless category\textgreater] papers; insufficient alignment will be rejected.  

\textbf{Format constraints:} plain text only. Each of Title, Abstract, Keywords on a single line; exact prefixes ``Title:'', ``Abstract:'', ``Keywords:'' with two spaces before ``Abstract:'' and ``Keywords.''  

\subsection{Prompt for Refinement with Feedback}
You have generated: \textless generated\_paper\textgreater.  
Similarity analysis: \textless feedback\textgreater. 

\textbf{Task: Revise} the paper to align more closely with [\textless category\textgreater] while maintaining originality:  
-- retain Title--Abstract--Keywords format;  
-- use examples of similar papers as guidance;  
-- extract 15--20 key terms \emph{from the revised abstract} and place them immediately after the abstract (two spaces before ``Keywords:''). 

\textbf{Format requirements:} same as above (plain text, single line per field, exact prefixes).

\subsection{Prompt for Text Modification}
\textbf{Anchor paper (Category: \{categories[anchor\_category]\}):} \{anchor\_text\}  

\textbf{Similar papers:} \{similar\_texts\}  

\textbf{Task:} Modify the anchor paper so that:  
-- it clearly remains in its category;  
-- it is more distinctive from similar papers;  
-- key medical concepts and methodologies are preserved, but the research focus or terminology is varied;  
-- output format: Title--Abstract--Keywords only, no explanations.

\subsection{Prompt for Edge Analysis}
\textbf{Anchor paper (Category: \{categories[anchor\_category]\}):} \{anchor\_text\}  
\textbf{Candidate papers:} \{similar\_texts\} 

\textbf{Task:} Decide which candidates should connect to the anchor based on:  
-- clinical methodology similarity;  
-- shared research domains or therapeutic targets;  
-- conceptual/medical relationships (e.g., biomarkers, mechanisms).  

\textbf{Edge rules:}  
-- connect same-category papers with strong methodological proximity;  
-- connect across categories only with strong overlap;  
-- be selective.  

\textbf{Output format:} Paper 1: CONNECT/REMOVE; Paper 2: CONNECT/REMOVE; ...

\section{Datasets}\label{datasets}

We conduct experiments on three widely used text-attributed citation graph datasets: \textbf{Cora}, \textbf{PubMed}, and \textbf{Arxiv}. The Arxiv dataset here is a sampled subset of the original OGB-Arxiv. 
Table~\ref{tab:datasets} summarizes their statistics.

\begin{table}[h]
\centering
\renewcommand{\arraystretch}{0.3}
\caption{Statistics of the datasets.}
\label{tab:datasets}
\begin{tabular}{lcccc}
\toprule
Dataset & \#Nodes & \#Edges & \#Features & \#Classes \\
\midrule
Cora   & 2,708  & 5,278  & 1,433 & 7  \\
PubMed & 19,717 & 44,324 & 384   & 3  \\
Arxiv  & 2,107  & 1,758  & 128   & 10 \\
\bottomrule
\end{tabular}
\end{table}

\section{Additional Experimental Results}

The full robustness results of GAT and GraphSAGE are reported in Appendix (Tables~\ref{GAT_robustness} and~\ref{GraphSAGE_robustness}), as a complement to the GCN results shown in the main paper (Table~\ref{robustness}). GRCN and RS-GNN are designed with GCN backbones, hence they are not applicable to GAT, and their GAT results are not reported in Table~\ref{GAT_robustness}.

\begin{figure*}[h]
  \centering
  \includegraphics[width=\linewidth]{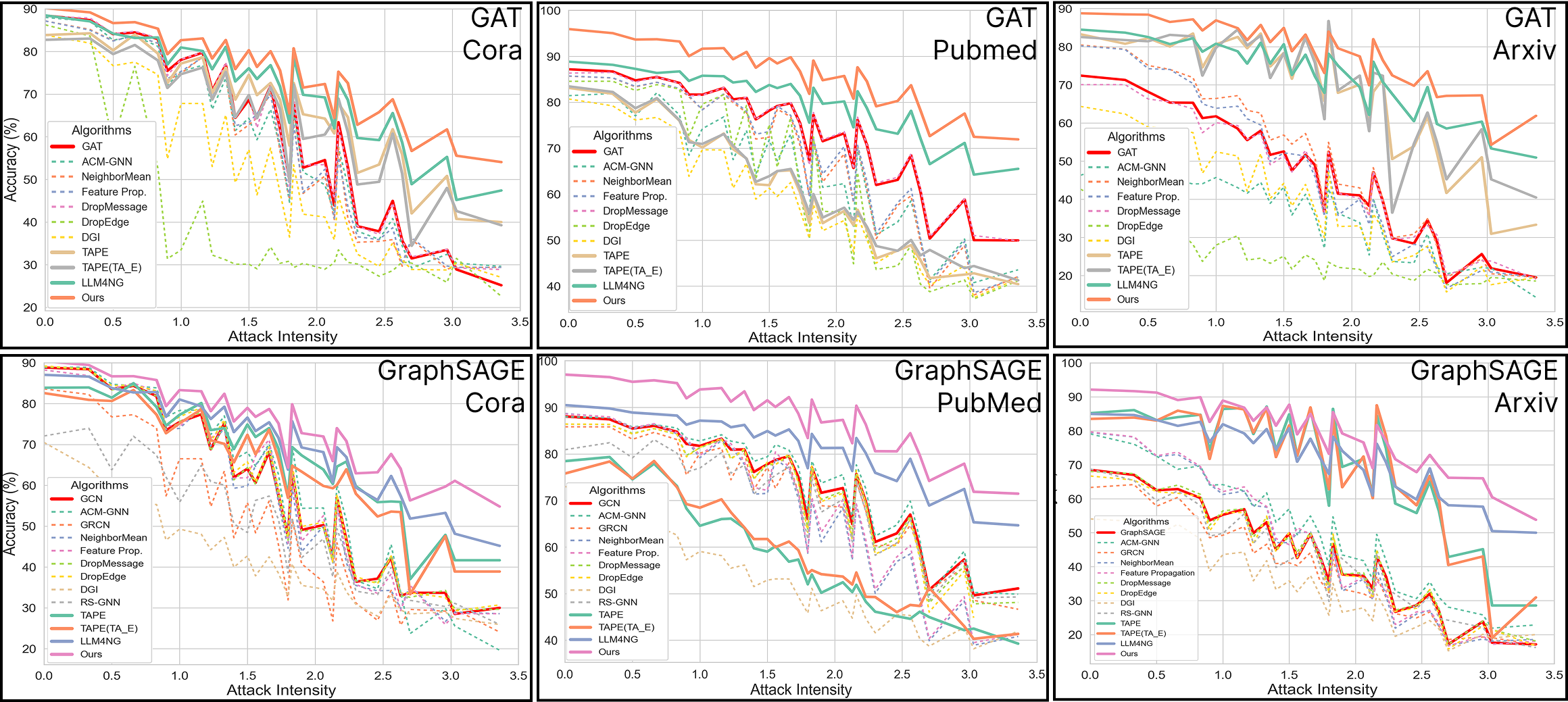}
  \caption{Performance curves of GAT and GraphSAGE on Cora, PubMed, and Arxiv.}
  \label{supply_performance_curves}
\end{figure*}

For completeness, we also report the performance curves of GAT and GraphSAGE across Cora, PubMed, and Arxiv datasets in Figure~\ref{supply_performance_curves}. These results complement the GCN robustness curves shown in the main paper (Figure~6).

\begin{table}[h]
\centering
\caption{GAT robustness.}
\label{GAT_robustness}
\Large
\renewcommand{\arraystretch}{1.2}
\resizebox{\linewidth}{!}{%
\begin{tabular}{l|ccc|ccc|ccc|ccc}
\toprule
\multirow{2}{*}{Algorithm} 
 & \multicolumn{3}{c|}{Clean Acc} 
 & \multicolumn{3}{c|}{Worst Acc} 
 & \multicolumn{3}{c|}{Avg Acc} 
 & \multicolumn{3}{c}{Norm-AUC} \\
 
 & Cora & Pub. & Arxiv 
 & Cora & Pub. & Arxiv 
 & Cora & Pub. & Arxiv 
 & Cora & Pub. & Arxiv \\
\midrule
GAT         & 88.43 & 87.19 & 72.45 & 77.72  & 82.64 & 62.32  & 83.68 & 85.03 & 67.91 & 0.84 & 0.85 & 0.68 \\
DropEdge    & 86.25 & 84.57 & 42.66 & 42.68 & 75.36 & 26.28 & 64.58 & 81.00 & 34.05 & 0.66 & 0.82 & 0.34 \\
DropMsg     & 88.28 & 86.35 & 70.07 & 77.00 & 82.43 & 59.49 & 83.45 & 84.57 & 65.60 & 0.84 & 0.85 & 0.66 \\
Feat. Prop. & 87.16 & 85.73 & 80.24 & 76.14 & 82.18 & 67.29 & 82.13 & 83.95 & 74.35 & 0.83 & 0.84 & 0.75 \\
Nbr. Mean   & 87.16 & 85.69 & 80.48 & 75.86 & 81.99 & 68.28 & 82.07 & 83.84 & 74.98 & 0.83 & 0.84 & 0.76 \\
ACM-GNN     & 87.99 & 81.44 & 46.32 & 75.01 & 71.73 & 41.08 & 82.25 & 77.20 & 46.06 & 0.83 & 0.78 & 0.46 \\
DGI         & 83.85 & 80.73 & 64.37 & 60.48 & 67.11 & 49.71 & 74.11 & 74.92 & 57.98 & 0.76 & 0.76 & 0.59 \\
TAPE        & 83.86 & 83.13 & 83.30 & 75.41 & 74.03 & 75.95 & 80.19 & 78.65 & 80.38 & 0.81 & 0.79 & 0.81 \\
TA\_E       & 82.74 & 83.48 & 82.57 & 73.98 & 73.72 & 74.67 & 79.01 & 78.93 & 79.74 & 0.80 & 0.79 & 0.80 \\
LLM4NG      & 88.47 & 88.85 & 84.51 & 78.51 & 85.09 & 79.21 & 83.96 & 87.15 & 82.19 & 0.84 & 0.87 & 0.8 \\
Ours        & \cellcolor{gray!25}\textbf{90.17} & \cellcolor{gray!25}\textbf{95.94} & \cellcolor{gray!25}\textbf{88.79} & \cellcolor{gray!25}\textbf{81.25} & \cellcolor{gray!25}\textbf{90.96} & \cellcolor{gray!25}\textbf{84.84} & \cellcolor{gray!25}\textbf{86.23} & \cellcolor{gray!25}\textbf{93.64} & \cellcolor{gray!25}\textbf{87.36} & \cellcolor{gray!25}\textbf{0.87} & \cellcolor{gray!25}\textbf{0.94} & \cellcolor{gray!25}\textbf{0.88} \\
\bottomrule
\end{tabular}}
\end{table}

\begin{table}[h]
\centering
\caption{GraphSAGE robustness.}
\label{GraphSAGE_robustness}
\Large
\renewcommand{\arraystretch}{1.2}
\resizebox{\linewidth}{!}{%
\begin{tabular}{l|ccc|ccc|ccc|ccc}
\toprule
\multirow{2}{*}{Algorithm} 
 & \multicolumn{3}{c|}{Clean Acc} 
 & \multicolumn{3}{c|}{Worst Acc} 
 & \multicolumn{3}{c|}{Avg Acc} 
 & \multicolumn{3}{c}{Norm-AUC} \\
 
 & Cora & Pub. & Arxiv 
 & Cora & Pub. & Arxiv 
 & Cora & Pub. & Arxiv 
 & Cora & Pub. & Arxiv \\
\midrule
GraphSAGE        & 88.84 & 88.06 & 68.51 & 76.20 & 83.07 & 55.95 & 83.28 & 85.72 & 62.70 & 0.84 & 0.86 & 0.63 \\
DropEdge         & 89.24 & 86.41 & 66.70 & 77.14 & 80.91 & 53.33 & 83.92 & 84.04 & 60.98 & 0.85 & 0.84 & 0.62 \\
DropMsg          & 89.06 & 86.41 & 68.03 & 75.70 & 81.12 & 55.16 & 83.12 & 84.14 & 62.33 & 0.84 & 0.85 & 0.63 \\
Feat. Prop.      & 88.25 & 88.59 & 79.71 & 77.30 & 84.14 & 66.96 & 83.35 & 86.32 & 73.58 & 0.84 & 0.86 & 0.74 \\
Nbr. Mean        & 88.25 & 88.68 & 79.48 & 76.53 & 83.78 & 66.50 & 83.08 & 86.17 & 73.28 & 0.84 & 0.86 & 0.74 \\
ACM-GNN          & 88.91 & 87.98 & 79.09 & 78.65 & 84.15 & 65.21 & 84.45 & 86.14 & 72.57 & 0.85 & 0.86 & 0.73 \\
DGI              & 70.50 & 73.01 & 54.06 & 48.84 & 59.14 & 42.10 & 59.78 & 66.35 & 48.82 & 0.60 & 0.67 & 0.50 \\
RS-GNN           & 72.09 & 80.91 & 68.77 & 61.77 & 77.03 & 57.21 & 67.66 & 79.59 & 63.09 & 0.65 & 0.80 & 0.63 \\
GRCN             & 83.62 & 85.83 & 63.33 & 62.40 & 77.73 & 51.51 & 74.72 & 82.46 & 58.08 & 0.76 & 0.83 & 0.59 \\
TAPE             & 83.90 & 78.46 & 85.22 & 76.94 & 70.69 & 75.83 & 81.07 & 74.97 & 81.98 & 0.82 & 0.75 & 0.83 \\
TA\_E            & 82.59 & 75.80 & 83.52 & 74.80 & 70.66 & 74.14 & 79.57 & 74.32 & 80.68 & 0.80 & 0.75 & 0.82 \\
LLM4NG           & 87.06 & 90.48 & 84.99 & 78.28 & 86.53 & 77.88 & 83.29 & 88.82 & 82.15 & 0.84 & 0.89 & 0.83 \\
Ours             & \cellcolor{gray!25}\textbf{90.39} & \cellcolor{gray!25}\textbf{97.04} & \cellcolor{gray!25}\textbf{92.16} & \cellcolor{gray!25}\textbf{81.24} & \cellcolor{gray!25}\textbf{92.85} & \cellcolor{gray!25}\textbf{84.24} & \cellcolor{gray!25}\textbf{86.34} & \cellcolor{gray!25}\textbf{95.23} & \cellcolor{gray!25}\textbf{89.26} & \cellcolor{gray!25}\textbf{0.87} & \cellcolor{gray!25}\textbf{0.96} & \cellcolor{gray!25}\textbf{0.90} \\
\bottomrule
\end{tabular}}
\end{table}

\end{document}